%% file: arxiv.tex
\documentclass{article} 
\usepackage{iclr2027_conference,times}

\input{math_commands.tex}

\usepackage{hyperref}
\usepackage{url}

\usepackage{url}            
\usepackage{booktabs}       
\usepackage{amsfonts}       
\usepackage{nicefrac}       
\usepackage{xcolor}         
\usepackage{amsmath}
\usepackage{mathtools}
\usepackage{amsthm}
\newtheorem{definition}{Definition}
\newtheorem{theorem}{Theorem}

\title{Revealing Pathway Interactions in \\Network Computations with \\Complete Path Contribution}

\author{%
    Shenghua Liu$^{*,\dagger}$\\
    Department of Physics\\
    Stanford University \\
    \And
    Youssef Faragalla$^*$\\
    Neurosciences Interdepartmental Program\\
    Stanford University \\
    \And
    Surya Ganguli\\
    Department of Applied Physics\\
    Stanford University \\
    \And
    Stephen A. Baccus$^\dagger$\\
    Department of Neurobiology\\
    Stanford University \\
    \thanks{$^*$These authors contributed equally to this work.}
    \thanks{$\dagger$Correspondence to: \texttt{sliu24@stanford.edu} or \texttt{baccus@stanford.edu}.}
}
\let\oldfootnotemark\footnotemark
\renewcommand{\thefootnote}{}
\renewcommand{\footnotemark}{}

\iclrfinalcopy 
\begin{document}

\maketitle

\renewcommand{\thefootnote}{\arabic{footnote}}
\let\footnotemark\oldfootnotemark
\setcounter{footnote}{0}

\begin{abstract}
Networks pass signals through many layers of interconnected units, so explaining what they compute means tracing how information travels from input to output and how pathways interact. Most existing approaches analyze only pairwise interactions between components such as units, layers or attention heads. Even when these measurements are combined post hoc into a circuit, they do not capture higher-order interactions along full causal chains (paths) from input to output, and so cannot address how exponentially many paths interact to construct network computation. This leaves a large set of mechanistic questions unanswerable, for example whether inhibitory pathways veto excitatory ones, and how the two combine to yield network decisions. To address this gap, we introduce Complete Path Contribution, a method to decompose the network output into mathematically complete contributions of individual paths to study path-level interactions directly and at scale. Partial summation then induces definitions of contributions for any circuit at any granularity level, with completeness guaranteed at every level. We further present an efficient algorithm that identifies the $k$ strongest paths without examining all of them, with runtime empirically linear in $k$ rather than exponential in depth. By applying Path Contribution to benchmark image classification networks, we quantify excitatory-inhibitory interactions and show how they collectively construct network decisions at a level of detail previously unmeasurable. Furthermore, in an interpretable model of the mouse retina, it reveals how dynamic receptive fields emerge from excitatory-inhibitory interaction motifs among pathways between different cell types in response to time-varying natural stimuli. Overall, Path Contribution reveals how a network's causal pathways interact at scale to transform input to output.
\end{abstract}

\section{Understanding network computations: from neurons to circuits}

Understanding neural network computations matters for advancing neuroscience and AI \citep{kriegeskorte2015deep, richards2019deep}, for model distillation \citep{hinton2015distilling}, and for AI safety \citep{bereska2024mechanistic}. Artificial and biological networks both compute in a distributed manner, transforming the input through many stages of nonlinear operations \citep{dicarlo2012does, lecun2015deep}, often through the combined effect of excitatory (positively contributing) and inhibitory (negatively contributing) units and pathways~\citep{isaacson2011inhibition, maheswaranathanInterpretingRetinalNeural2023}. Understanding a network therefore requires understanding not only its individual units, but also how they coordinate in interacting pathways to transform input to output.

Parallel pathways carrying distinct signals are central to the visual system \citep{nassi2009parallel}. In primates, the parvocellular pathway is higher resolution, slower, chromatically sensitive, and less adaptive to contrast, whereas the magnocellular pathway is lower resolution, faster, chromatically insensitive, and more adaptive \citep{derrington1984spatial, kaplan1986primate, solomon2004profound}. In cortex, the ventral stream is associated with object discrimination and the dorsal stream with estimating object position and trajectory \citep{Ungerleider1982,goodale1992separate}.

Interactions between excitation and inhibition have long been known to underlie visual computation. Edge enhancement in the retina arises not from edge-specific excitation, but because inhibition through parallel pathways cancels excitation in uniform regions away from edges \citep{ratliff1959responses, kuffler1953discharge, barlow1953summation}. Similarly, object motion sensitivity, which flags moving objects while rejecting background eye movements, arises because all retinal motion generates excitation, but inhibition vetoes it wherever motion is coincident across the visual field \citep{olveczkySegregationObjectBackground2003, baccusRetinalCircuitThat2008}. That this computation arises only two synapses after edge enhancement, yet was discovered decades later, shows how challenging it is to connect interacting pathways to computation. As features are computed through serial layers, canceling paths become difficult to find because their net effect is zero. Identifying such mechanisms requires measuring the contributions of individual pathways, including those that cancel, rather than only the net effects of units.

Excitation-inhibition interactions also underlie computations in artificial networks. In InceptionV1, a car detector is excited by wheel features below and window features above, and inhibited by the same features in the opposite positions; selectivity arises from subtracting a parallel pathway rather than from excitation alone \citep{olahZoomIntroductionCircuits2020}. In GPT-2 small, some attention heads copy the correct name to the output while others suppress the token the model is about to predict, apparently calibrating its confidence \citep{wangInterpretabilityWildCircuit2022}. Such cancellation can hide components from analysis. In language models, ablating one head often triggers compensation by others, so a component's net effect can be near zero while its direct contribution is large \citep{mcgrath2023hydra}. Excitation and inhibition are therefore properties of paths, not units.

\subsection{Two orthogonal axes of explanation}

A full explanation of network computation is composed of two orthogonal axes. The parallel (or disjunctive) axis typically involves first specifying a cut of the model's computational graph (such as neurons or weights in a layer). Then, the explanation offers an interpretable account of ``what is happening at this stage of the network'', such as features represented or causally important neurons. Next, the second axis extends this explanation across depth, which is the serial (or conjunctive) direction of explanation. Traditionally, one applies the method to many layers separately and observes the evolution of the features or causal information contained, and sometimes pairwise interactions between adjacent layers, but this does not constitute a full explanation because it does not answer how representations or actions compose across many layers of the network to drive the output. This higher-order composition happens in a combinatorial number of ways growing exponentially with depth, which makes it an intractable problem so far. 

\subsection{The landscape of interpretability tools}

The first stage of explanation is understanding how a single unit (neuron or component) computes. A canonical framework from neuroscience characterizes a unit by how the input drives it, its \emph{receptive field} (RF), and how it drives the output, its \emph{projective field} (PF)~\citep{lehkyNetworkModelShapefromshading1988a}. RF methods include spike-triggered averaging~\citep{deboerTriggeredCorrelation1968}, instantaneous receptive fields (IRFs)~\citep{maheswaranathanInterpretingRetinalNeural2023}, most exciting inputs~\citep{walkerInceptionLoopsDiscover2019}, and sparse autoencoders~\citep{bricken2023monosemanticity}. PF methods include gradient-based attribution~\citep{simonyanDeepConvolutionalNetworks2014,anconaBetterUnderstandingGradientbased2018}.

Other methods measure a unit's total causal effect on the output, such as the raw gradient~\citep{simonyanDeepConvolutionalNetworks2014}, SmoothGrad~\citep{smilkovSmoothGradRemovingNoise2017}, and Integrated Gradients~\citep{sundararajanAxiomaticAttributionDeep2017} for inputs, and LRP~\citep{bachPixelWiseExplanationsNonLinear2015}, DeepLIFT~\citep{shrikumarLearningImportantFeatures2017}, and Deep SHAP~\citep{lundbergUnifiedApproachInterpreting2017} for hidden neurons. Causal interventions~\citep{vigInvestigatingGenderBias2020,geigerCausalAbstractionsNeural2021} instead compare the original network with a modified one, so they do not explain a unit's PF in the unaltered network. None of these methods show how the RF and PF compose. A related line of work does express a unit's contribution as a composition of its RF and PF, including activation times gradient~\citep{shrikumarLearningImportantFeatures2017}, neuron conductance~\citep{dhamdhereHowImportantNeuron2018,shrikumarComputationallyEfficientMeasures2018}, its application to retinal models~\citep{tanakaDeepLearningMechanistic2019a,maheswaranathanInterpretingRetinalNeural2023}, and contribution decomposition (CODEC)~\citep{melanderCausalInterpretationNeural2026}, which further analyzes population motifs of contributions within a layer.

However, fully explaining a neuron's RF requires more than relating its input to its activation. We must examine the network in between, since the transformation arises from many interacting parallel pathways. Likewise, explaining its PF requires analyzing how the pathways from the neuron to the output interact. A satisfactory method must therefore quantify each pathway's role in a neuron's computation, with guarantees that pathway-level scores fully account for the neuron's effect, and that neuron-level effects in turn aggregate to fully account for the network output.

Existing circuit-level interpretability methods do not offer a satisfactory solution to this problem. One line of related work here is circuit discovery~\citep{conmyAutomatedCircuitDiscovery2023a,syedAttributionPatchingOutperforms2024,marksSparseFeatureCircuits2024,ameisen2025circuit} and model pruning~\citep{hatefiPruningExplainingRevisited2024,hatefiAttributionGuidedPruningInsight2026}. These methods often identify the most important units first via causal interventions~\citep{mengLocatingEditingFactual2022} or attribution~\citep{kramarAtPEfficientScalable2024}, and then compose them into a sparse circuit. These circuits may be more interpretable than the full model, and in some cases tailored to semantic interpretation~\citep{geigerCausalAbstractionsNeural2021}, but they do not quantify the contributions of individual paths. A small literature directly quantifies the importance of paths~\citep{luInfluencePathsCharacterizing2020,luInfluencePatternsExplaining2021,duGeneralizingBackpropagationGradientBased2023,goldowsky-dillLocalizingModelBehavior2023,wangInterpretabilityWildCircuit2022}, but these works do not advance a notion of hierarchical completeness for the paths of a neuron and, in turn, the neurons of a network.

\subsection{A unified framework for analyzing network computations at the unit and circuit levels at scale}

\begin{figure}[ht]
    \centering
    \includegraphics[width=0.8\linewidth]{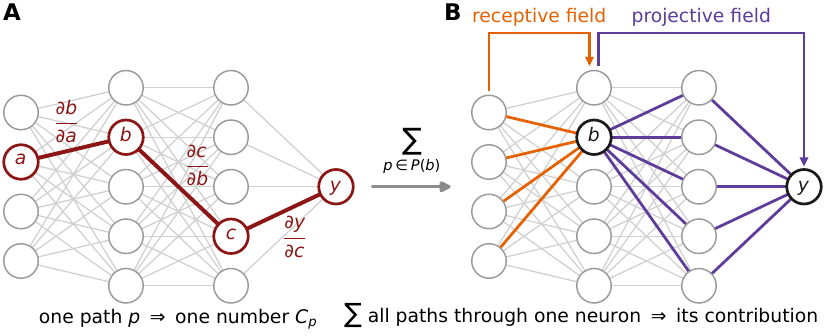}
    \caption{(A) The contribution of a path is the chained gradient along its edges integrated over a straight line path in input space. (B) A neuron's contribution is a composition of its receptive and projective fields. The RF and PF both involve many parallel pathways, and they compose individually into full paths from input to output.}
    \label{fig:schematic}
\end{figure}

To address this gap, we introduce a principled and unified framework, Complete Path Contribution, to quantify the causal contribution of each neuron, component, subnetwork (circuit), and full input-output pathway and analyze them at scale (\Figref{fig:schematic}). Our main contributions are as follows:
\paragraph{One algorithm for neurons, components, subnetworks, and paths.} Unlike previous methods that first find important components in each layer and then compose them into a circuit post hoc, we propose a single algorithm that directly measures the contribution of each path, which naturally induces the contribution of any subnetwork as a sum over its paths. In particular, the contribution of an input pixel and a hidden neuron recover Integrated Gradients and neuron conductance~\citep{dhamdhereHowImportantNeuron2018} respectively, and the contribution of a convolutional channel recovers CODEC~\citep{melanderCausalInterpretationNeural2026}.

\paragraph{Hierarchical completeness.} Under our framework, a coarse component's contribution is guaranteed to be an exact sum of the contributions of its finer components (e.g., paths through a neuron, neurons in a convolutional channel), and the sum over any cut of the model's computational graph is guaranteed to equal the output difference with respect to a baseline. See Appendix~\ref{sec:subnetwork} for the full definition of subnetwork contributions and the proof of hierarchical completeness.

\paragraph{An efficient algorithm for finding the $k$ strongest paths.} A full explanation of network computations involves characterizing a number of paths that is exponential in the number of layers. We tackle this intractable problem by  devising an efficient algorithm to identify the $k$ strongest paths without exhaustive enumeration. Instead, the runtime is empirically linear in the number of paths requested.
Overall, this work enables a unified view of component-based and circuit-based explanations of network computation, and makes an initial effort towards a full explanation of network computations at the path level via an efficient top-$k$ algorithm.

\section{Path Contribution at Scale}


\subsection{The contribution of a path}

The Path Contribution framework starts with defining the contribution of a single path. Consider a feedforward neural network $f$ with input $\vx$, baseline $\vx'$, and scalar output $y=f(\vx)$. Let the input be layer $0$ and the output be layer $L$. Let $\tA$ be the activations of the network, with $\etA^l_{i_l}$ being the activation of neuron $i_l$ in layer $l$, so $\tA^0=\vx$ and $\tA^L=y$. Let $\mathcal{P}$ be the set of all paths from the input to the output, where a path $p\in\mathcal{P}$ is a sequence of neurons $(i_0, i_1, \dots, i_L)$ with $i_l$ being a neuron in layer $l$. Let $\alpha\in[0,1]$ parameterize a straight-line path from the baseline to the input, so that $\tA^0(\alpha)=\vx'+\alpha(\vx-\vx')$. 

\begin{definition}[Contribution of a path]
\label{def:path-contribution}
We define path contribution by starting at the output difference, invoking the completeness of Integrated Gradients, and expanding the chain rule sum:
\begin{align}
\Delta y
= \sum_{i_0} \mathrm{IG}^0_{i_0} 
&= \sum_{i_0} (\evx_{i_0} - \evx'_{i_0}) \int_0^1 \frac{\partial f\big(\vx' + \alpha(\vx-\vx')\big)}{\partial \evx_{i_0}}\, d\alpha \\
&= \sum_{i_0} \int_0^1 \frac{\partial y}{\partial \etA^0_{i_0}} \frac{\partial \etA^0_{i_0}}{\partial \alpha}\, d\alpha \\
&= \sum_{i_0} \int_0^1 \sum_{i_1,\dots,i_L} \frac{\partial \etA^L_{i_L}}{\partial \etA^{L-1}_{i_{L-1}}} \cdots \frac{\partial \etA^1_{i_1}}{\partial \etA^0_{i_0}} \frac{\partial \etA^0_{i_0}}{\partial \alpha}\, d\alpha \\
&= \sum_{i_0,i_1,\dots,i_L} {{\int_0^1 \frac{\partial \etA^L_{i_L}}{\partial \etA^{L-1}_{i_{L-1}}} \cdots \frac{\partial \etA^1_{i_1}}{\partial \etA^0_{i_0}} \frac{\partial \etA^0_{i_0}}{\partial \alpha}\, d\alpha}}\label{eq:path-completeness}
\end{align}
The summand is our definition of the contribution of a path $(i_0, i_1, \dots, i_L)$:
\begin{equation}
\etC_{i_0, i_1, \dots, i_L} \coloneqq \int_0^1 \frac{\partial \etA^L_{i_L}}{\partial \etA^{L-1}_{i_{L-1}}} \cdots \frac{\partial \etA^1_{i_1}}{\partial \etA^0_{i_0}} \frac{\partial \etA^0_{i_0}}{\partial \alpha}\, d\alpha .
\label{eq:path-contribution}
\end{equation}
\end{definition}
We recover Integrated Gradients by summing over all paths that end at the same input pixel, and neuron conductance/contribution $\sum_{i_l} \int_0^1 \frac{\partial y}{\partial \etA^l_{i_l}} \frac{\partial \etA^l_{i_l}}{\partial \alpha}\, d\alpha$ by summing over all paths that end at the same hidden neuron. Furthermore, path contribution can be taken over a distribution of inputs, for example to measure the average contribution of a path over a dataset. Completeness still holds due to the linearity of expectations.

This definition of path contribution is mathematically equivalent to influence paths~\citep{luInfluencePathsCharacterizing2020}. However, we extend this definition to a hierarchical framework that naturally defines the contribution of any subnetwork, with a guarantee of completeness at every level.

\subsection{\texorpdfstring{Top-$k$}{Top-k} algorithm}

The main challenge of fully explaining network computations rests in the intractability of the compositional interactions between components across layers: explaining what happens at every layer of a network does not answer how they interact across layers. At the level of paths, our framework turns this challenge into a single parallel aggregation of an exponential number of paths. Even though characterizing that exponential set remains prohibitive in general, we devise an efficient algorithm to identify the $k$ paths with the highest absolute contribution without exhaustive enumeration.

Without integration and for a single stimulus, the path contribution is purely a product of the Jacobians along the path. Taking the logarithm of the absolute value of the contribution, the problem of finding the top-$k$ paths reduces to finding the $k$ longest paths in a directed acyclic graph (DAG), which can be solved in time polynomial in both $k$ and the network size~\citep{tziavelisAnykAlgorithmsEnumerating2025}. \citet{luInfluencePatternsExplaining2021} remarked that an algorithm with such worst case guarantees is impossible due to the summation over the integration steps and inputs, but an algorithm that is efficient in practice is still possible. We present such an algorithm that, within the range we tested, scales up to a 12-layer MLP with 128 neurons per layer ($784\times128^{12}\approx10^{28}$ paths) in reasonable time on an M3 Pro MacBook Pro. 

The algorithm is a best-first branch and bound search, analogous to the A* algorithm, which uses a heuristic bound to find an optimal path without exhaustively searching the graph. The intuition is as follows. Let a \emph{prefix} $(i_0,\dots,i_l)$ denote the set of all its completions. We now organize the path set as a tree, each node being a prefix. The root is empty, which stands for all paths, and the leaves are individual paths. For every node we compute an upper bound on $|\etC_p|$ over all its completions. We keep a priority queue ordered by bound, and always expand the most promising node. When a node is expanded into complete paths, each is pushed with its exact $|\etC_p|$ in place of a bound. When a complete path is at the top of the queue, it is guaranteed to be stronger than any path not yet seen, since its exact contribution outranks every bound still in the queue. The first $k$ paths to be popped from the queue are the top-$k$ paths. A complete description of the algorithm and its pseudocode is given in Appendix~\ref{sec:top-k-algorithm}.

\section{Neurons and Paths: A Toy Interpretability Case Study}

\begin{figure}[ht]
    \centering
    \includegraphics[width=0.9\textwidth]{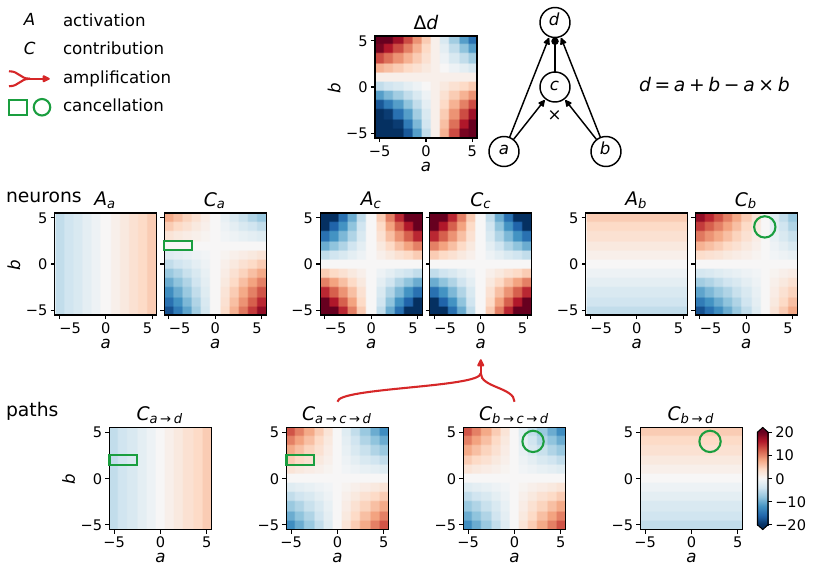}
    \caption{A toy model with two inputs computes $d=a+b-a\times b$. Each heatmap shows the relevant quantity as a function of inputs. Contributions are computed with a zero baseline, so $\Delta d=d$ is the output, shown at the top. The middle row shows the activation and contribution of the three neurons. The bottom row shows the contribution of each of the four paths. Due to hierarchical completeness, $\etC_{ad}+\etC_{acd}=\etC_{a}$, $\etC_{acd}+\etC_{bcd}=\etC_{c}$, $\etC_{bcd}+\etC_{bd}=\etC_{b}$, and $\etC_{a}+\etC_{b}=\Delta d$. Observe cancellations indicated by the green boxes and green circles, which are not discoverable at the neuron level.}
    \label{fig:toy}
\end{figure}

In \Figref{fig:toy}, we consider a toy model that approximates a continuous exclusive-or (XOR) function and examine what conclusions we would draw from a neuron-based vs. path-based approach regarding its computation. With activation analysis, we would be able to characterize the RFs of $c$ and $d$ separately, examining which inputs excite them. Because their RFs are of opposite sign, one may conclude that their RFs differ by an inhibitory connection, but nonetheless cannot deduce that such a connection causally connects the two. Analyzing neuron contributions additionally allows for causal conclusions. Because the activation and contribution of $c$ are of opposite sign, the net causal effect of $c$ on $d$ is negative whenever $c$ is active, which yields a causal statement that $c$ inhibits $d$ as its projective field. However, this analysis misses interactions at the pathway level. In the input region indicated by the green boxes, an inhibitory direct effect from $a$ to $d$ is canceled by an excitatory effect along the path mediated by $c$. Analysis of $a$'s contribution would conclude that it is not causally important at that input, even though it is active. This misses the E-I interaction in $a$'s PF that results in a net zero effect. Therefore, a full explanation of network computation necessitates path-level explanations.

The toy model also illuminates how Path Contribution relates to causal interventions. Path patching~\citep{goldowsky-dillLocalizingModelBehavior2023} simulates the output change if a specific path's effect was replaced by that from a counterfactual input, here the zero baseline. In the framework of Shapley values~\citep{rozemberczkiShapleyValueMachine2022}, a full characterization of a path's causal effect among $k$ total paths involves measuring its patching effect in $2^{k-1}$ coalitions (whether each remaining path is patched or not; we will call these \emph{contexts}), the mean of this score distribution being the Shapley value. Path Contribution is a causal measure based on accumulating gradients (which are inherently causal) instead of patching, but it coincides with path patching on linear networks. Furthermore, in this nonlinear toy model, Path Contribution coincides with the Shapley values, while path patching from the full model doesn't, motivating Path Contribution as a useful representative statistic of the full causal score distribution that enjoys the completeness property. See Appendix~\ref{sec:path-patching} for details.

\section{Pathways Interact to Construct the Projective Field}

\begin{figure}[!htbp]
    \centering
    \includegraphics[width=0.9\textwidth]{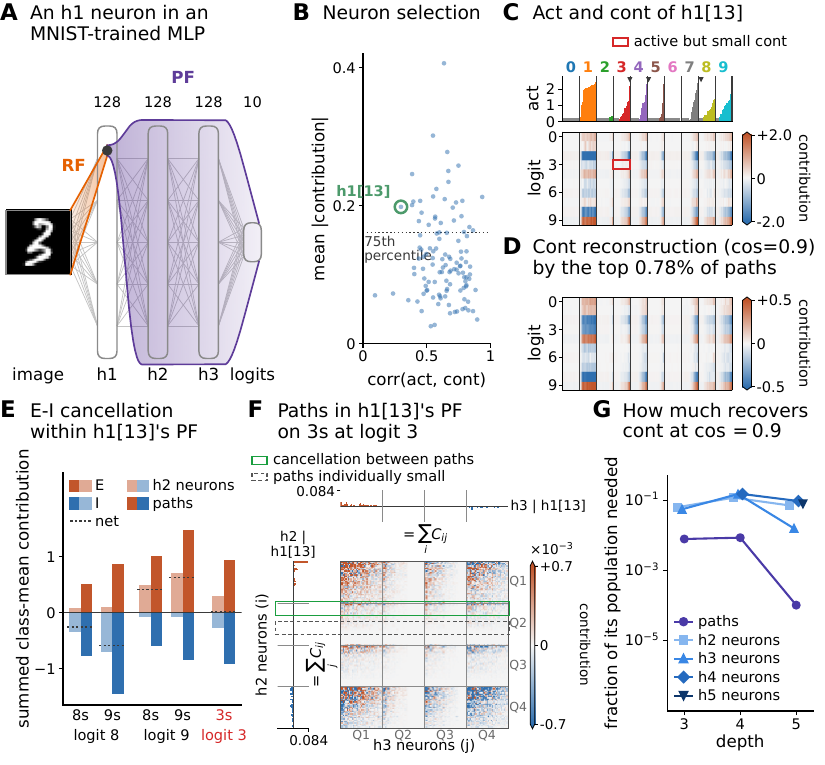}
    \caption{Analysis of pathway interactions in a ``silent'' neuron's PF in an MNIST-trained MLP. A silent neuron is one that is active but contributes little to the output. (A) The model: A 3-layer MLP with 128 neurons per layer trained on MNIST. (B) The silent neuron selected by a set of criteria: activity---the neuron is active on at least 30\% of the validation set, importance---its mean absolute contribution over the validation set to the correct logit is in the top quartile, and silence---a small correlation between its activation and contribution. (C) The activation and contribution of the selected neuron h1[13] on each of the 100 (input image class, output logit) scenario, over 20 random images of each class. Contribution is computed with a zero baseline at 64 integration steps. h1[13] has an approximately linear PF: one dominant pattern over the logits, scaled by the activation. However, a nonlinear feature is indicated by the scenario in the red box, where on images of 3s it barely contributes at all to logit 3 despite being consistently inhibitory for other images. (D) A small fraction of paths in the neuron's PF reconstructs the contribution pattern over scenarios with high accuracy. (E) Total E and I effects in h1[13]'s PF at the neuron (which includes h2 and h3 neurons) and path levels, in five settings. Neuron contributions are mostly decided by the output logit, and are mostly all E or all I. Paths, however, exhibit a high level of cancellation resulting in the same residual sum. On images of 3 taken to the 3 logit (3s@3), E and I effects almost cancel perfectly at both the neuron and path levels. (F) The full path contribution matrix conditioned on h1[13], with paths in a row summing to an h2 neuron's contribution and columns to h3 neurons. Sorted by neuron contribution into quartiles, and within each quartile sorted by the variance. Many h2 and h3 neurons have no net effect, yet path contribution reveals that it is due to large cancellation between paths in some cases (green box) and paths being individually weak in others (gray). An enlarged version of this plot is in \Figref{fig:silent-neuron-matrix}. (G) A smaller fraction of paths is needed to reconstruct the contribution properties of h1[13] compared to downstream neurons, suggesting more sparsity at the path level.}
    \label{fig:silent-neuron}
\end{figure}

A neuron's contribution is composed of its RF and PF. Contributions are often sparser than activations, because an active neuron (RF activated) could have a small downstream gradient (PF not activated) (see \citep{maheswaranathanInterpretingRetinalNeural2023} Fig.~6 and \citep{melanderCausalInterpretationNeural2026} Fig.~2). In such cases, it is then unknown whether the downstream effect being small is due to individual pathways not crossing threshold, or whether the pathways are individually strong but cancel. We use MNIST-trained MLPs as an illuminating example to investigate this question. Throughout this paper, all MLPs are ReLU networks trained with PyTorch default parameters, using Adam at $10^{-3}$ learning rate, over 5 epochs, at 128 batch size, and reaching over 95\% test accuracy. In \Figref{fig:silent-neuron}, we study a neuron in the first hidden layer (h1) of a 3-layer MLP that has high activation but low contribution on images of 3 taken to the logit 3 (3s@3). The neuron is selected by a set of criteria: a high level of activity, a substantial contribution to the correct logit, and a low activation-contribution correlation (panel B). The PF of this neuron, h1[13], is revealed by its activations and contributions to all 10 logits across 20 images of each class (panel C). It is roughly linear, showing a consistent pattern over the logits across inputs that is scaled by activation. However, we focus on the nonlinear feature, which is why the neuron is active on 3s@3 but not contributing much as opposed to consistently being inhibitory for other image classes. We then asked why its net contribution is small, are downstream pathways in its PF individually weak and do not cross threshold, or are they strong but cancel? Neuron-level contribution methods cannot answer this question, as they only measure the net effect. We analyze the $128\times128=16384$ paths in its PF, each path implicitly summing over inputs. In this context, the neuron-level effect of h2 and h3 neurons refer to all the paths that pass through both h1[13] and the selected h2 or h3 neuron, which is in fact an order-2 (edge) level effect viewed in the whole model. Plotting the total E and I effects in h1[13]'s downstream neurons and paths across stimuli and logits (panel E) reveals that the small net contribution results from canceling pathways similar in magnitude to other scenarios where the net effect is large and positive (logit 9) and large and negative (logit 8). In addition, neuron-level net effects (h2 neurons in the plot) are largely single-signed, whereas path contributions have larger E and I totals that cancel to the same residual sum. Therefore, neuron-level contribution analyses overlook the large amounts of E-I interaction at the path level. To visualize this effect, panel F shows the contribution of all paths in h1[13]'s PF, going through a select h2 neuron and h3 neuron. The green and gray boxes indicate regions where an h2 neuron's total effect (marginal) is small, yet the decomposition into its paths over h3 neurons (a row in the region) shows significant E and I activity that largely cancel in one case and individually small E and I activity in the other. Furthermore, paths offer a sparser description of the PF than neurons. Panel G shows that about 100 times fewer paths are needed by fraction to reconstruct the PF to mean cosine similarity 0.9, whose visualization is shown in panel D. Overall, this analysis demonstrates an example where path contribution produces answers to questions at the pathway level that cannot be answered with neuron-level analyses.

\begin{figure*}[!tp]
\centering
\includegraphics[width=\textwidth,height=0.7\textheight,keepaspectratio]{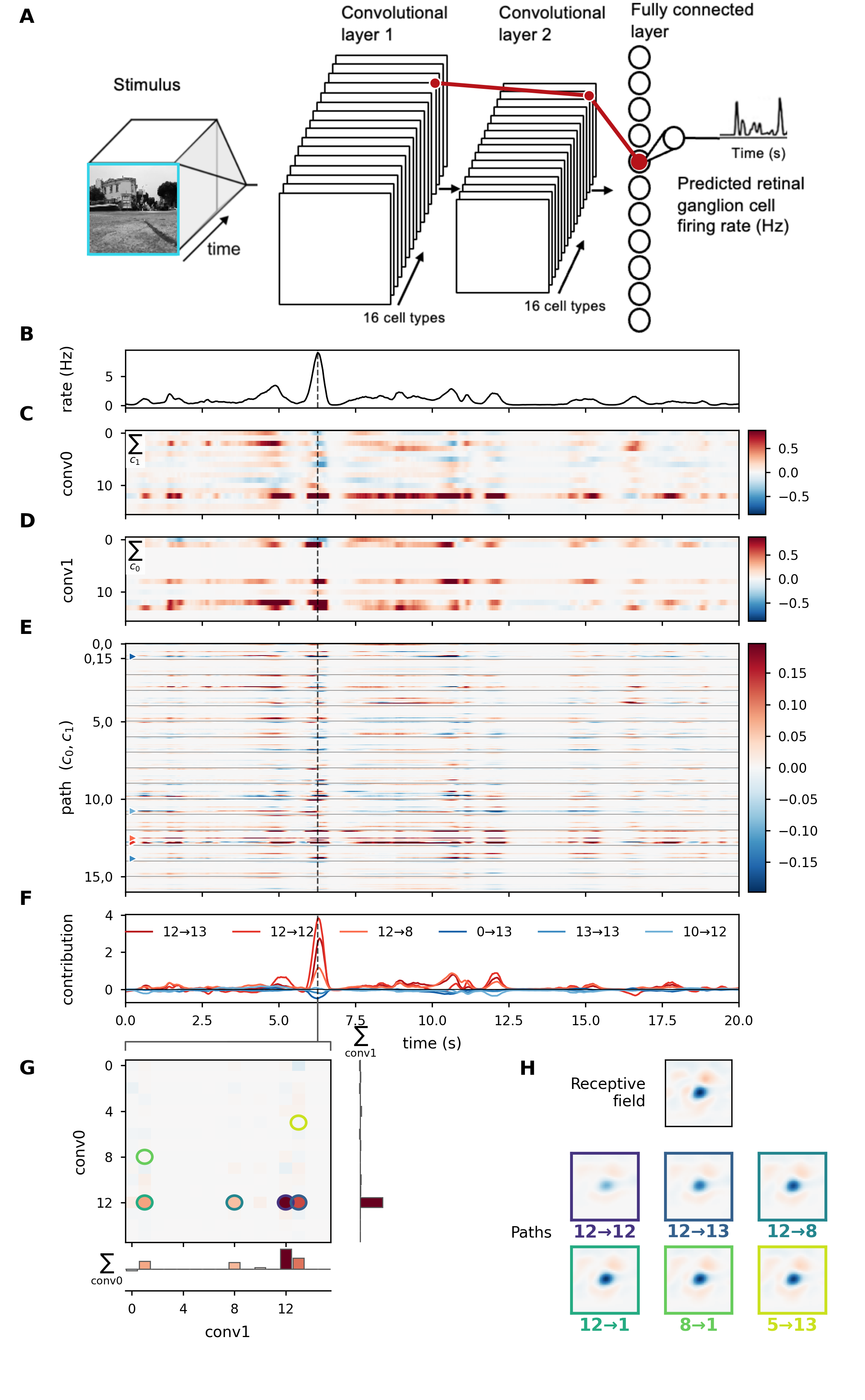}
\caption{Path contributions to activity and receptive fields of a retinal ganglion cell.
(A) Schematic diagram of a 3-layer CNN fit to mouse retinal ganglion cell firing rate responses. A $30$-frame (100 x 100) window of a natural movie
 drives two layers of $16$ channels each, read
out through a fully connected layer to the rates of all $36$ simultaneously
recorded cells. Paths are computed for each hidden unit and summed over one pair $(c_0,c_1)$ of channels, one path of the $16\times16=256$ through the network. The path shown in red,
$12\rightarrow13$, is the strongest by mean contribution and the first trace in
panel F. 
(B) Predicted firing rate of one cell,
$r=0.57$ against the recorded rate. The dashed line marks the firing peak used in
panels G and H.
(C, D) Contributions of the $16$ conv0 and conv1 units at each time point, obtained by summing the path matrix over the other layer's
index. These values are channel-wise contributions for a given layer \citep{melanderCausalInterpretationNeural2026}.
(E) All 256 path contributions $P_t[c_0,c_1]$ grouped by
conv0 channel; panels C and D are its two marginals. Arrows mark
the paths traced in panel F. 
(F) The three strongest excitatory and  inhibitory
paths, ranked by mean contribution over the displayed interval.
(G) The path matrix at the marked time point, flanked by the two marginals
of panels C and D; each marginal is labelled with the index it
sums over. Circles mark the paths whose filters are added in panel H.
(H) The instantaneous receptive field (IRF) \citep{maheswaranathanInterpretingRetinalNeural2023} assembled from those paths. Colored borders match paths in panel G. Three paths of $256$ sum to r = $0.95$ with the IRF.}
\label{fig:retina-fig1}
\end{figure*}

\begin{figure*}[!tp]
\centering
\includegraphics[width=\textwidth,height=0.7\textheight,keepaspectratio]{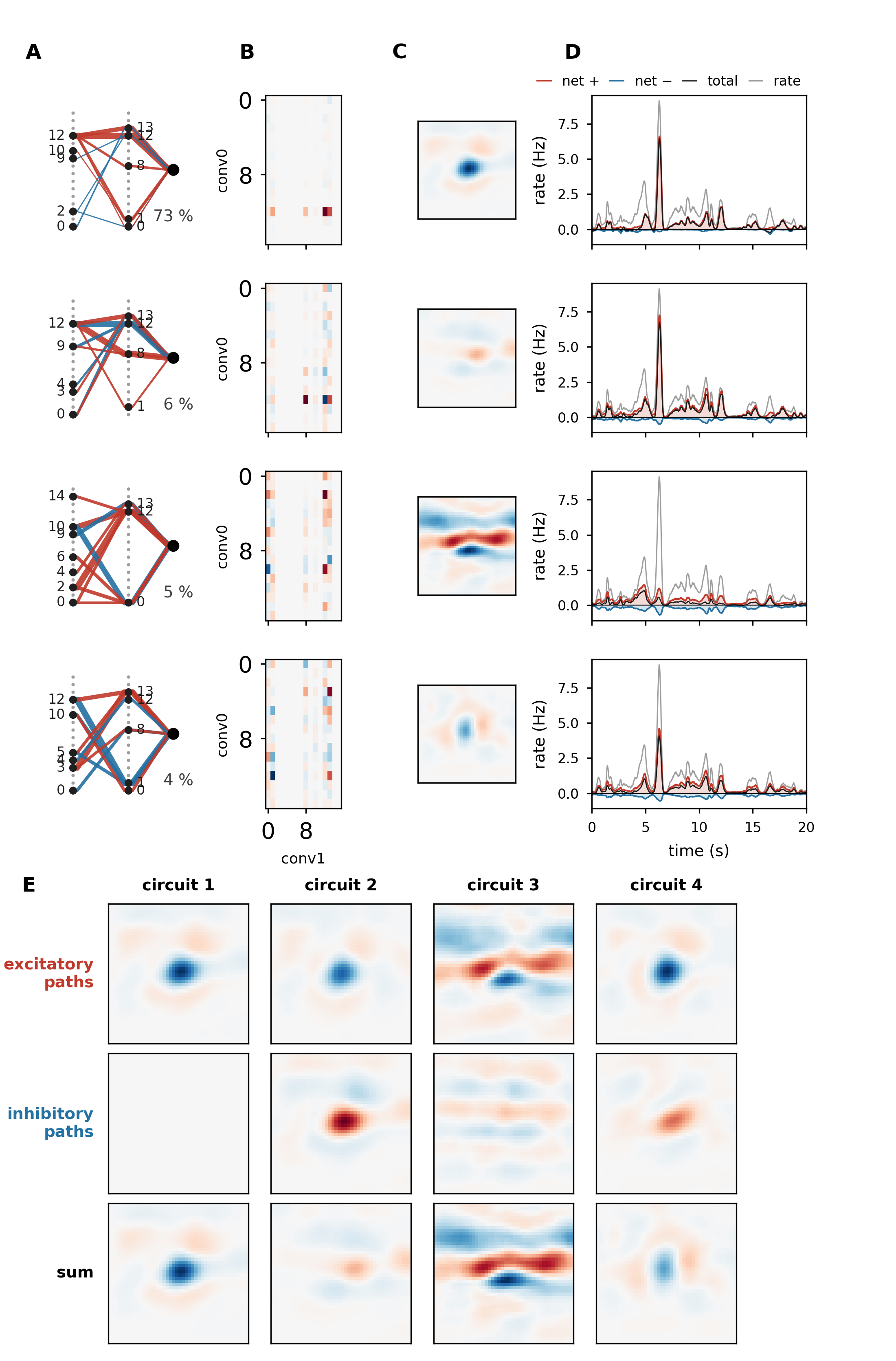}
\caption{Circuits formed by paths acting in concert.
Principal components of the $256\times T$ matrix of path contributions, taken
over the whole clip.
{(A)}~Each circuit drawn over the channel graph, along with the percentage of variance it accounts for. A circuit's paths are those contributing above $5\%$ of its
maximum ($15$, $45$, $59$ and $70$ paths, respectively). One line per path, its
width the magnitude of that path's average contribution in the circuit and its color the
sign. Line width is the path
contribution, not the layer-to-layer weight.
{(B)}~Path contributions of the same circuits shown as a $16\times16$ matrix.
{(C)}~The average receptive field that the circuit implements. 
{(D)}~Contribution of each circuit over time. Shown at each time point are summed contributions for the paths in panel A labelled in red (net positive) or blue (net negative), and the total contribution (black) along with the firing rate (gray). All contributions sum to the firing rate minus the response to the baseline, which is $0.05$~Hz for this cell.
{(E)}~Receptive field contributions for circuit components that are either excitatory or inhibitory, on average, summed over a long stretch of data. Each component has positive and negative values because local increases and decreases in intensity can drive either excitation or inhibition.}
\label{fig:retina-fig2}
\end{figure*}

Path Contribution is applicable to any feedforward architecture. In Appendix~\ref{sec:vit}, we apply Path Contribution to pairs of attention heads and residual streams in different layers of a ViT-B/16 model trained on ImageNet-1k~\citep{dosovitskiyImageWorth16x162020}. We find that contributions concentrate in different types of head-stream interactions depending on depth; we also find coordinated E-I actions between head pairs, showing the prevalence of pathway interactions across architectures.

\section{Decomposing cell-type effects in a retinal model}

Previous work used a generalizable and mechanistically interpretable three-layer CNN model of the retina (\Figref{fig:retina-fig1}, panel A) to identify how individual model neurons drive the output by measuring their contributions \citep{maheswaranathanInterpretingRetinalNeural2023}. Although this approach can be applied at any layer, it treats each layer in isolation: it distinguishes the effects of different units within a layer on the retinal output, but does not reveal how units act in series through the network. We therefore used path contribution analysis to examine causal effects from input to output. In a model of mouse retinal ganglion cells with 16 channels (cell types) in each of the two hidden layers, each ganglion cell's response is carried by 256 paths, one for each pair of channels $(c_0, c_1)$ (panel E). Marginalizing the path contributions $P_t[c_0,c_1]$ over $c_1$ or $c_0$ recovers the contribution of each conv0 or conv1 cell type, respectively (panels C, D and G), so the paths reveal the effects that combine to produce each cell type's contribution. For a given cell type, the paths passing through it carried both excitation and inhibition, which partially canceled to yield its net contribution (panels E and F). The strongest inhibitory paths converged on the same conv1 channels targeted by the strongest excitatory paths (panel F). Path contributions thus localize where the cancellation occurs, which the cell-type marginals alone cannot reveal. After summing over space, some paths changed sign over time, whereas others remained consistently excitatory or inhibitory (panel E). Inhibitory paths became transiently active at the same moment the excitatory paths peaked, suggesting that inhibition shapes or sharpens the firing event rather than acting as tonic suppression (panel F). Finally, because path contributions are complete, we could ask how many paths are needed to reconstruct the firing rate and the instantaneous receptive field (panels G and H).

We next examined how paths act in concert by applying principal component analysis (PCA) to the $256 \times T$ matrix of path contributions over the clip (\Figref{fig:retina-fig2}). The leading components identified a small number of motifs, or ``circuits,'' each coordinating excitation and inhibition across both hidden layers (panels A and B). For the ganglion cell shown, the first four circuits accounted for $88\%$ of the variance (panel A). Each circuit implemented a distinct receptive field (panel C), and excitation and inhibition played a different role in each (panel E). These interactions are inaccessible to standard contribution analysis, which assigns a single net value to each unit, and they reveal how paths coordinate to shape the model's response. 

\section{Conclusion}
We introduce Complete Path Contribution, a unified framework to compute the contribution of any subnetwork, including paths, edges, neurons, and components, with the same algorithm. Furthermore, this framework guarantees hierarchical completeness, where the contributions of a subnetwork's constituents sum to the subnetwork contribution itself, holding across all levels from individual paths to the network output difference. This guarantee sets a standard scale to compare the amount of cancellation happening at different levels, which other methods without this guarantee do not have. To analyze the exponential number of paths at scale, we devise an efficient algorithm that finds the $k$ strongest paths in empirically linear time, taking a first step towards a full understanding of network computation at the path level. On benchmark image classification models and an interpretable model of the mouse retina, we demonstrate that path contribution offers answers to mechanistic questions at the path level that neuron-level methods cannot answer. We show in an MNIST-trained MLP how cancellation between a neuron's downstream pathways leads to its minimal net contribution despite being activated. We show E-I interaction motifs between attention heads across layers of a ViT that vary and flip sign across inputs. We also show that a single cell type-level path in the retinal model approximates the receptive field of a ganglion cell, and that path-level E-I motifs describe network action that dynamically vary over time. Overall, path contribution reveals important mechanisms of network computation that are invisible to previous methods, and holds promise for understanding how the brain and artificial neural networks compute.

\subsection*{AI use statement}

In this work, we used generative AI tools to aid or polish writing. We used these tools to review mechanics, improve flow, and to generate sound bites that facilitate writing. We used generative AI for retrieval and discovery. We used these tools to search literature on a particular topic and summarize them. When important, we read the original source and added it to the references if applicable. We used these tools for ideation and execution. We used them to generate ideas about the top-$k$ algorithm, which we then manually followed through and verified. We used these tools for execution: we used them to write and refactor code, which we verified by reading it and testing on simple cases. We used these tools to draft sections of the paper. For example, we used them to generate pseudocode for the top-$k$ algorithm that we then manually verified and modified for better presentation. We also used these tools for proving mathematical claims, and also to generate figures whose numbers we also verified. We first derived all claims by hand, and then used the tools to generate a streamlined presentation which we verified. We have not used these tools to generate synthetic datasets.

\subsection*{Ethics statement}

No human subjects or personally identifiable data are involved, and the image classification experiments use public benchmarks and publicly released checkpoints. The mouse retinal data was collected under a protocol approved by the Stanford Institutional Animal Care and Use Committee and in accordance with NIH guidelines. The authors declare no competing interests.

\subsection*{Reproducibility statement}

We provide precise definitions of path contribution and contribution of subnetworks, how to compute them in practice, along with a detailed explanation and pseudocode for the top-$k$ algorithm. Experiments use MLPs trained with a standard pipeline with a detailed description of hyperparameters, a trained ViT-B available from Torchvision, and an interpretable model of the mouse retina trained on electrophysiology recordings. The implementation, trained retinal and image classification models, the mouse retinal recordings and the analysis pipeline will be released in a public repository on publication.

\subsubsection*{Acknowledgments}
This work was supported by grants from the National Eye Institute (NEI): R01EY022933,
R01EY025087 and P30EY026877 (SAB). S.G. thanks the Simons Foundation Collaboration on the
Physics of Learning and Neural Computation and a Schmidt Sciences Polymath Award for support. Y.F. was supported by the Stanford HAI-Data Science Scholars Program.

\bibliography{references}
\bibliographystyle{iclr2027_conference}

\appendix

\section{The Contribution of Any Subnetwork}
\label{sec:subnetwork}

Because the computational graph of a feedforward network is a directed acyclic graph (DAG) $\gG=(\gV,\gE)$, any subnetwork (circuit) $\mathcal{H}\subseteq\gG$ is also a DAG. In this graph setting, neurons are nodes and weights are edges. Let $(l,i_l)\in\gV_l$ denote the $i_l$-th neuron in layer $l$, an element of the set of layer-$l$ neurons $\gV_l$. Let $(l,i_l)\rightarrow(l+1,i_{l+1})\in\gE_{l,l+1}$ denote the edge from neuron $(l,i_l)$ to neuron $(l+1,i_{l+1})$, an element of the set of edges $\gE_{l,l+1}$ from layer $l$ to layer $l+1$. 

\begin{definition}[Contribution of a subnetwork]
\label{def:subnetwork-contribution}
A subnetwork $\mathcal{H}$ is given by the allowed neurons and edges in each layer:
\begin{equation}
\mathcal{H}=\big(\gV^{\mathcal{H}}_0,\ \gE^{\mathcal{H}}_{0,1},\ \gV^{\mathcal{H}}_1,\ \dots,\ \gE^{\mathcal{H}}_{L-1,L},\ \gV^{\mathcal{H}}_L\big),
\qquad \gV^{\mathcal{H}}_l\subseteq\gV_l,\quad \gE^{\mathcal{H}}_{l,l+1}\subseteq\gE_{l,l+1}.
\end{equation}
For example, a single neuron is the special case in which every layer but one keeps its full set of neurons and edges. The allowed neurons and edges in each layer induce masks on that layer's neurons and edges, $\vm^l\in\{0,1\}^{|\gV_l|}$ and $\mM^{l,l+1}\in\{0,1\}^{|\gV_{l+1}|\times|\gV_l|}$, with
\begin{equation}
\evm^l_{i}=\1\big[(l,i)\in\gV^{\mathcal{H}}_l\big],
\qquad
\emM^{l,l+1}_{j,i}=\1\big[(l,i)\rightarrow(l+1,j)\in\gE^{\mathcal{H}}_{l,l+1}\big].
\end{equation}
Let $\mathcal{P}(\mathcal{H})$ be the set of paths that pass through $\mathcal{H}$, that is, whose every neuron and edge is allowed. The contribution of $\mathcal{H}$ is defined as the sum of the contributions of these paths:
\begin{equation}
\etC(\mathcal{H})=\sum_{p\in\mathcal{P}(\mathcal{H})}\etC_p
=\sum_{i_0,\dots,i_L}\etC_{i_0,\dots,i_L}\prod_{l=0}^{L}\evm^l_{i_l}\prod_{l=0}^{L-1}\emM^{l,l+1}_{i_{l+1},i_l}.
\label{eq:subnetwork-contribution}
\end{equation}
\end{definition}

\begin{table}[t]
\caption{Special cases of Definition~\ref{def:subnetwork-contribution}. Layers not listed keep their full sets of neurons and edges.}
\label{tab:subnetwork-examples}
\begin{center}
\begin{tabular}{lll}
\toprule
Subnetwork $\mathcal{H}$ & Masks & $\etC(\mathcal{H})$ \\
\midrule
Input pixel $i_0$ & $\vm^0=\ve_{i_0}$ & Integrated Gradients \\
Neuron $(l,j)$ & $\vm^l=\ve_j$ & Neuron conductance \\
Component $\sS\subseteq\gV_l$ & $\vm^l=\1_{\sS}$ & Component contribution, e.g., conv channel \\
Edge $(l,j)\rightarrow(l+1,k)$ & $\mM^{l,l+1}=\ve_k\ve_j^\top$ & Edge (weight) contribution \\
Path $(i_0,\dots,i_L)$ & $\vm^l=\ve_{i_l}$ for every $l$ & Path contribution \\
\bottomrule
\end{tabular}
\end{center}
\end{table}

In practice, with $N_\alpha$ integration steps, we compute the contribution of a path with $N_\alpha$ backward passes, one per step. The contribution of any subnetwork also takes only $N_\alpha$ backward passes: at each step, we put the masks where they belong in the backward pass, multiplying the gradient at layer $l$ by $\vm^l$ and the weights between layers $l$ and $l+1$ by $\mM^{l,l+1}$. If the Jacobians between all pairs of constrained layers are saved, this process can be parallelized, as in CODEC~\citep{melanderCausalInterpretationNeural2026}.

\begin{theorem}[Hierarchical completeness]
\label{thm:hierarchical-completeness}
Let $\mathcal{H}$ be a subnetwork, and let $\mathcal{H}_1,\dots,\mathcal{H}_K$ be subnetworks whose path sets partition that of $\mathcal{H}$, $\mathcal{P}(\mathcal{H})=\bigsqcup_{k=1}^{K}\mathcal{P}(\mathcal{H}_k)$. Then
\begin{equation}
\etC(\mathcal{H})=\sum_{k=1}^{K}\etC(\mathcal{H}_k).
\end{equation}
In particular, $\etC(\gG)=\Delta y$, so the contributions of any family of subnetworks whose path sets partition $\mathcal{P}$ sum to $\Delta y$.
\end{theorem}

\begin{proof}
By Definition~\ref{def:subnetwork-contribution}, $\etC(\mathcal{H})$ is the sum of $\etC_p$ over $p\in\mathcal{P}(\mathcal{H})$. Since the $\mathcal{P}(\mathcal{H}_k)$ partition $\mathcal{P}(\mathcal{H})$, 
\begin{equation*}
\etC(\mathcal{H})=\sum_{p\in\mathcal{P}(\mathcal{H})}\etC_p=\sum_{k=1}^{K}\sum_{p\in\mathcal{P}(\mathcal{H}_k)}\etC_p=\sum_{k=1}^{K}\etC(\mathcal{H}_k).
\end{equation*}
For $\mathcal{H}=\gG$, every mask is all ones, and $\etC(\gG)=\sum_{i_0,\dots,i_L}\etC_{i_0,\dots,i_L}=\Delta y$ by \Eqref{eq:path-completeness}.
\end{proof}

For example, write $\etC^{0,\dots,l}_{i_0,\dots,i_l}$ for the contribution of the subnetwork that fixes neuron $i_0$ in layer $0$, \dots, neuron $i_l$ in layer $l$, and keeps every other layer full. Applying Theorem~\ref{thm:hierarchical-completeness} at each level gives
\begin{equation}
\Delta y
=\sum_{i_0} \etC^{0}_{i_0}
=\sum_{i_0}\sum_{i_1} \etC^{0,1}_{i_0,i_1}
=\sum_{i_0}\sum_{i_1}\sum_{i_2} \etC^{0,1,2}_{i_0,i_1,i_2}
=\cdots
=\sum_{i_0,\dots,i_L}\etC_{i_0,\dots,i_L}.
\end{equation}
The output difference splits into the input pixels, as in Integrated Gradients; each pixel $i_0$ splits into its outgoing edges into the first hidden layer; each pixel-neuron pair $(i_0,i_1)$ splits into the three-neuron paths $(i_0,i_1,i_2)$ that extend it; and so on down to full paths.

\section{Choosing Baselines and Integration Steps}
\label{sec:baseline-steps}

There are two key parameters when applying Path Contribution: the baseline and the number of integration steps. For integration steps, we want to choose a sufficient number that the contribution of important paths is converged and the overall completeness error is small. For the baseline, we note that different baselines are asking different questions. Path Contribution fundamentally measures a counterfactual effect---the accumulated infinitesimal chained-gradient contributions along the line from the baseline to the input. For example, if we took an image as a baseline and integrate to another image, that would measure the contribution of a path to the output change as the input changes from one image to the other. If we took a null baseline instead, we would interpret the result as the contribution of the path to the output itself (minus a null output). Therefore, in general, the contribution of a path under different baselines need not agree, since it could play different roles in different input changes. However, it is reasonable to expect that under different null baselines, the resulting explanation remains consistent. Here, we give an empirical account of how key observations---cancellation amounts and patterns, path rankings, and path signs---stay consistent or vary across baselines and integration steps as an intuitive guide to choosing these parameters. Throughout this section we use the setting of \Figref{fig:silent-neuron}: the 16,384 paths from h1[13] through h2 and h3 to logit 3, on the 1,012 validation images of 3 (3s@3), with a black baseline and 64 integration steps unless stated otherwise.

\paragraph{Cancellation under different baselines.}

\begin{figure}[!htbp]
    \centering
    \includegraphics[width=\textwidth]{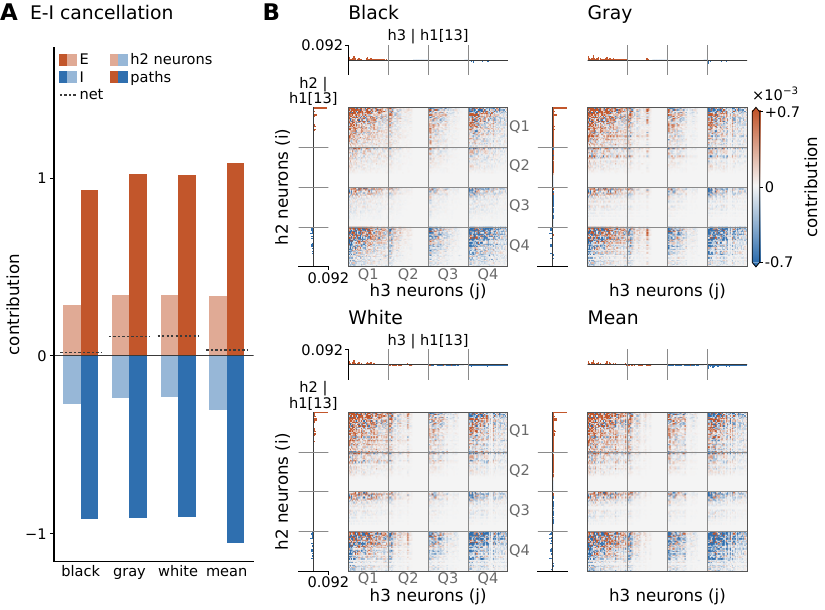}
    \caption{Interactions among h1[13]'s downstream pathways under four baselines, in the setting of \Figref{fig:silent-neuron} (3s@3, mean over all 1,012 validation images of 3, 64 integration steps). Black is the baseline of the main text; gray sets every pixel to $1/2$ and white to 1; mean is the mean of all MNIST validation images. (A) Total E and I through h2 neurons and through h2-h3 paths, with the net contribution as a dotted line, as in \Figref{fig:silent-neuron}E. E and I are similar under all four baselines, while the net contribution differs. (B) The class-mean path contribution matrix with its conditional marginals, as in \Figref{fig:silent-neuron}F. Rows and columns keep the order of the black baseline, and the four matrices and marginals share the same scale respectively.}
    \label{fig:baseline-matrices}
\end{figure}

Other than the default baseline (black), we take three additional null baselines: gray (halfway), white, and the mean of all MNIST validation images. As shown in \Figref{fig:baseline-matrices}A, the total E and I change little across baselines while the h1[13]'s net contribution differs slightly, suggesting that it could play slightly different roles in contrasting the actual input with the different baselines, while the mechanics of strong E-I cancellation remains. In \Figref{fig:baseline-matrices}B, we see that the E-I pattern stays highly consistent across baselines as well, suggesting that h1[13]'s PF circuit is robust across baselines.

\begin{figure}[!htbp]
    \centering
    \includegraphics[width=\textwidth]{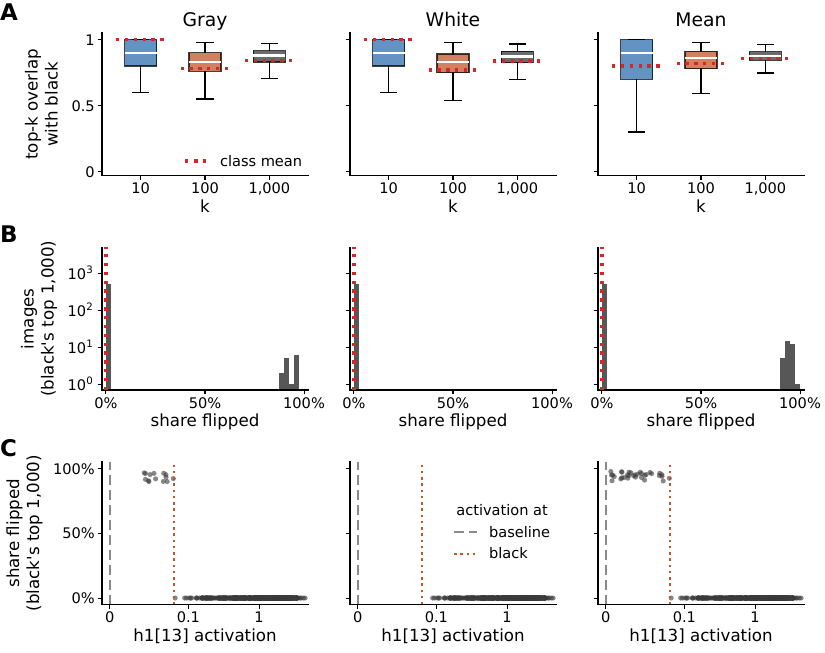}
    \caption{Path rankings and signs under three baselines, compared with the black baseline, in the setting of \Figref{fig:baseline-matrices}. Each column is one baseline. An image is left out of a column if its path contribution matrix is identically zero under that baseline (since ranking and sign flip become meaningless), which happens when h1[13] is inactive at every integration step; this leaves 541 (gray), 526 (white) and 560 (mean) of the 1,012 images. (A) Top-$k$ overlap with the black baseline, as a distribution over images, for $k=10$, 100 and 1,000 of the 16,384 paths. Boxes show quartiles and whiskers extend to 1.5 times the interquartile range; outliers are not drawn. Red dotted lines show the overlap of the class-mean paths. (B) Histogram over images of the share of the black baseline's top 1,000 paths whose contribution has the opposite sign under the column's baseline. The red dotted line marks the share for the class-mean matrices, which is zero. (C) The same share against the activation of h1[13] at the image. Vertical lines mark the activation of h1[13] at the column's baseline (dashed) and at the black baseline (dotted). Paths change sign only on images whose activation lies between the two.}
    \label{fig:baseline-rankings}
\end{figure}

\paragraph{Rankings and signs under different baselines.} We compare two rankings of paths by their top-$k$ overlap: the fraction of the $k$ paths with the largest absolute contribution under one baseline that are also among the top $k$ under the other, for $k=10$, 100 and 1,000. We use this metric over rank distance measures because of its intuitive picture of the ranking near the top when there are many very small or exactly zero paths that may confound other measures. As shown in \Figref{fig:baseline-rankings}A, the top path ranking is preserved (overlap >0.8 in general) for all three alternative baselines, whether on individual images or the class mean. As shown in panel B, on the vast majority of images and on the class mean, almost all paths keep their sign as baseline changes. On a select few images, almost all paths flip sign. This is investigated in panel C, which reveals that the reason they flip sign is because h1[13] is so weakly activated on the image that its activation sits between that on the black and the chosen baselines. Then, the two baselines are essentially integrating from opposite directions towards the middle, naturally yielding path contributions that flip sign. This is reflective of the fact that different baselines ask different questions, because integrating from a more exciting image to a less exciting one is asking how much a path contributes as its input goes down as opposed to go up. Furthermore, note that the white baseline does not have these minority images, since they get left out due to every path being exactly zero---a weakly activating neuron on the actual input is prone to getting exactly zero path contributions as it gets integrated towards a baseline.

\paragraph{Convergence of numerical integration.}

\begin{figure}[!htbp]
    \centering
    \includegraphics[width=\textwidth]{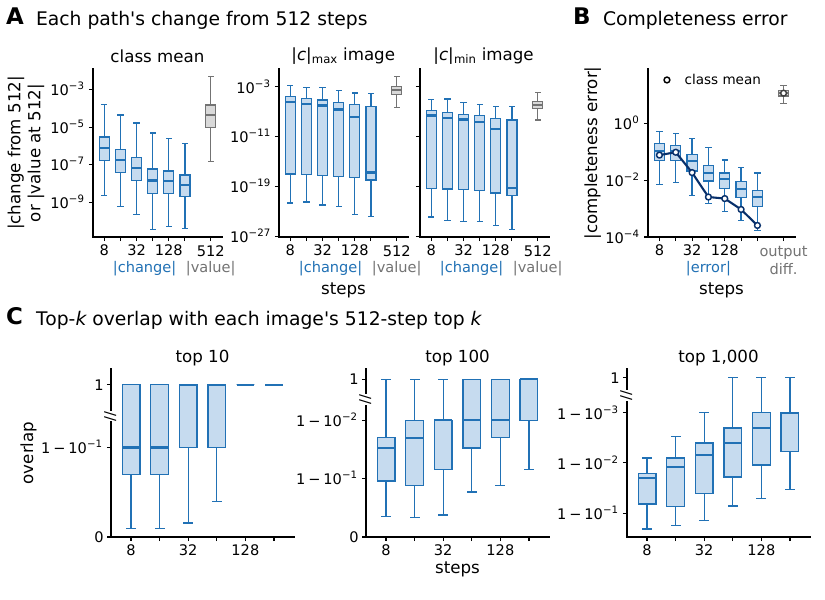}
    \caption{Convergence in the number of integration steps, in the setting of \Figref{fig:silent-neuron} (black baseline, all 1,012 validation images of 3). Boxes show quartiles and whiskers extend to 1.5 times the interquartile range; outliers are not drawn. (A) Blue: the absolute change of each path's contribution from its value at 512 steps, as a distribution over the 16,384 paths, at 8 to 256 steps. Gray: the absolute contribution at 512 steps, for reference. Left: the class-mean contributions. Middle and right: the two images with the largest and the smallest $|c|$ among the 561 images on which h1[13] is active, where $c$ is the contribution of h1[13] on the image at 512 steps. (B) Blue: the absolute completeness error---difference between the sum of all paths and the output change---, as a distribution over images, at 8 to 512 steps. Gray: the absolute output difference. Circles show the class mean. (C) Overlap between the top-$k$ paths of an image at each step count and its top-$k$ at 512 steps, as a distribution over images. The axis is logarithmic in one minus the overlap. Images whose path contributions are all exactly zero at a step count (179 at 8 steps and 38 at 16) are left out at that step count.}
    \label{fig:steps}
\end{figure}

We use the convergence of individual path contributions and the overall completeness error as a proxy of the approximation quality of a given number of integration steps. We show both in \Figref{fig:steps} for the setting in \Figref{fig:silent-neuron}. Since the contribution of a single path has no closed form, so we take its value at 512 steps as the reference and use the change from that value as a proxy for the error. We see in panel A that for the class mean and for individual images of the class, the error of individual path contributions become at least two orders of magnitude smaller than the typical path contribution at reasonable (such as 64, which we use in \Figref{fig:silent-neuron}) integration steps. Panel B shows that overall completeness error starts out already less than $1\%$ of the output difference and plunges quickly with integration steps. In panel C, all three top-$k$ overlaps quickly become high under reasonable integration steps, reaching over 0.9 on the vast majority of the images at 64 steps. Therefore, integration error is for all intents and purposes negligible at the \Figref{fig:silent-neuron} setting of 64 steps.

\section{Head-Head E-I Interaction Motifs in ViT-B}
\label{sec:vit}

\begin{figure}[ht]
\begin{center}
\includegraphics[width=\textwidth]{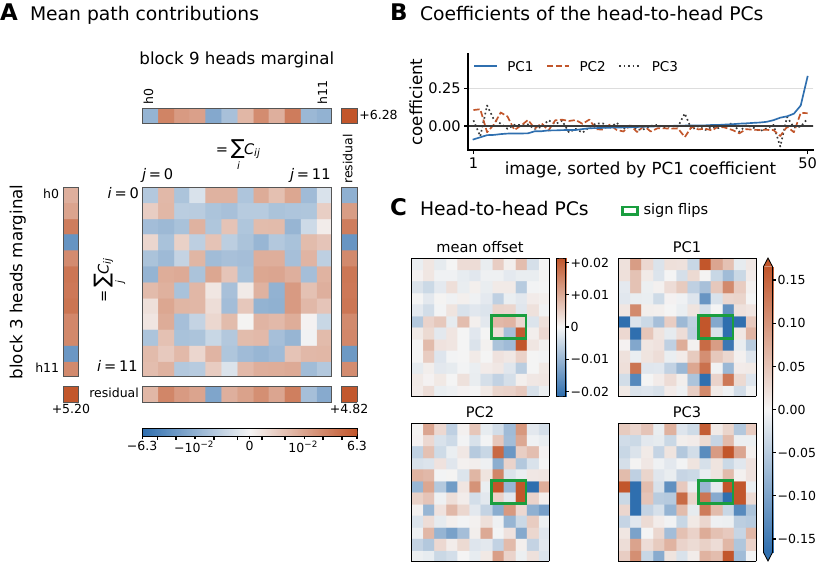}
\end{center}
\caption{Head-head interaction motifs in ViT-B trained on ImageNet. (A) Mean contributions of paths that pass through specified attention heads at block 3 and block 9, taken over the 50 validation images of the ``pill bottle'' class, using an ImageNet mean baseline and 512 integration steps. Paths that go through the residual stream in one layer and a head in the other dominate contribution. However, head-head contributions still reveal E-I interaction not visible at a single layer (rows and columns are sums over E and I paths contained in them). (B) PCA coefficients on the head-head contributions over 50 images sorted by PC1. Both PC2 and PC3's coefficients vary across inputs, and sometimes flip sign. (C) The mean offset and top three PCs of head-head contributions. }
\label{fig:vit}
\end{figure}

Transformers have become a dominant architecture in deep learning. Attention heads are a key component of the transformer architecture, and extensive work has been done to uncover the role of attention heads in a transformer's computation~\citep{elhage2021mathematical,wangInterpretabilityWildCircuit2022}. Path Contribution offers a general framework to analyze how head-head interactions contribute at any order (for example, pairwise, triplet, etc.) with guaranteed completeness. As an example, we analyze in \Figref{fig:vit} how pairwise interactions of attention heads contribute to the classification in a ViT-B/16 model trained on ImageNet-1k~\citep{dosovitskiyImageWorth16x162020} (from Torchvision).

For the baseline, we use the mean ImageNet image. For integration steps, we find that 512 steps is well-converged, with the completeness error of the class mean at 1.8\%, 0.9\% and 0.2\% of the output difference at 128, 256 and 512 steps, for the pill bottle class. For the other three classes we analyze, at 512 steps the completeness error of the class mean is 0.05\% for mailbag, 0.1\% for tape player and 1.1\% for convertible.

Between two chosen blocks, a path either passes through one of the 12 attention heads (h) or bypasses them through the residual stream, which we call the wire (w). This divides the paths into four parts (interaction types), named with the earlier block first: wire-wire (ww), head-wire (hw), wire-head (wh) and head-head (hh). In \Figref{fig:vit}, we find that even though the overall magnitude of head-head contributions are small compared to their coupling with the residual stream (panel A), their contributions naturally decompose into action motifs containing both excitatory and inhibitory paths that dynamically change magnitude and flip sign over different images of the same class (panel B). While we do not claim semantic interpretation of these motifs, their existence suggests the ubiquitous utilization of coordinated E-I interactions in network computation.

\begin{figure}[!htbp]
    \centering
    \includegraphics[width=\textwidth]{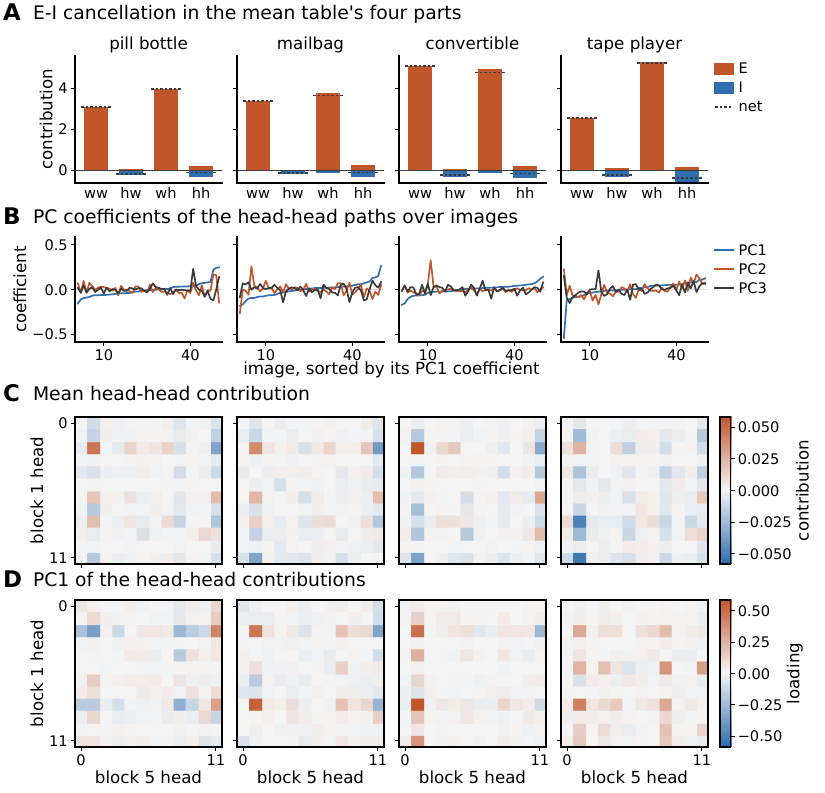}
    \caption{Head-head interactions in ViT-B across four classes, at blocks 1 and 5. Each column is one class, with its 50 validation images taken to the logit of the class. (A) Total E and I of the class-mean contributions in the four parts of the paths (types of head-stream interactions): ww, paths that bypass the heads of both blocks through the residual stream (the wire); hw, paths through a head of block 1 and the wire at block 5; wh, paths through the wire at block 1 and a head of block 5; hh, paths through a head of each block. The dotted line is the net contribution of the part. (B) Coefficients of the first three PCs of the head-head contributions over the 50 images, sorted by the PC1 coefficient. (C) The mean head-head contribution of each pair of heads. (D) The first PC of the head-head contributions. Each row shares one scale.}
    \label{fig:vit-classes}
\end{figure}

\begin{figure}[!htbp]
    \centering
    \includegraphics[width=0.8\textwidth]{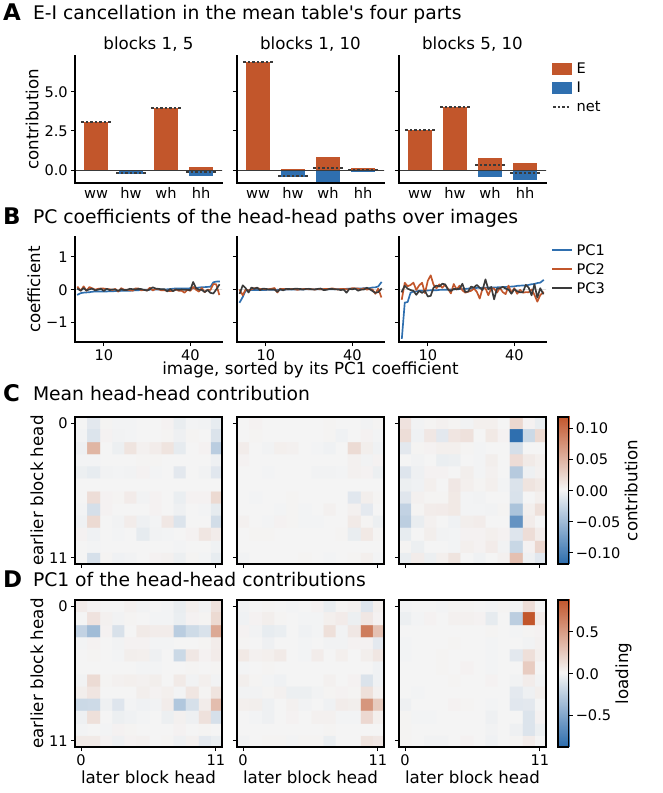}
    \caption{Head-head interactions in ViT-B across three pairs of blocks, for the 50 validation images of the pill bottle class. Each column is one pair of blocks; panels are as in \Figref{fig:vit-classes}, with the first letter of a part naming the earlier block. The four parts divide the output differently in each pair, the first PC and coefficient scales differ between pairs.}
    \label{fig:vit-pairs}
\end{figure}

We also analyze three other classes, mailbag, tape player, and convertible, chosen randomly, and three block-pair settings: blocks 1 and 5, 1 and 10, and 5 and 10. \Figref{fig:vit-classes} and \Figref{fig:vit-pairs} show E-I cancellation statistics and motifs for different classes on the same block 1-5 pair and different block pairs on the same pill bottle class, respectively.

Across classes (\Figref{fig:vit-classes}), the four interaction types divide the output in the same way. Wire-wire and wire-head carry most of it, and head-head has a small net contribution from the cancellation of E and I of similar size. In every class the PC coefficients of the head-head contributions change sign across images (B), and the mean head-head contributions and the first PC share their largest entries across the four classes (C, D). This suggests that part of the head-head computation is common to different classes, which is interesting to investigate in future work.

Across block pairs (\Figref{fig:vit-pairs}), the division changes. At blocks 1 and 5 wire-head is the largest, and at blocks 5 and 10 head-wire is; both consist of paths through a head of block 5. At blocks 1 and 10 wire-wire carries nearly the whole output. Paths through the heads of the middle block thus carry a large share of the output, while at blocks 1 and 10 most of the output bypasses the heads. E and I of similar size remain in the head-head part at blocks 1 and 5 and at blocks 5 and 10, and in the wire-head part at the two pairs that include block 10. The PC coefficients again change sign across images, with a larger spread at blocks 5 and 10, and the first PC differs significantly between the three pairs. For one class, the heads of different blocks therefore interact in different patterns, an interesting phenomenon to investigate in future work.

\section{Details of the \texorpdfstring{Top-$k$}{Top-k} Algorithm}
\label{sec:top-k-algorithm}

The goal of the top-$k$ algorithm is to return the $k$ paths with the largest $|\etC_p|$ efficiently. To compute the path contribution in practice, we discretize the integral in \Eqref{eq:path-contribution} with $N_\alpha$ steps $\alpha_1,\dots,\alpha_{N_\alpha}$ and weights $w_t$. For example, with the right Riemann sum, $w_t=1/N_\alpha$ and $\alpha_t=t/N_\alpha$. Writing $J^l_{i_l,i_{l-1}}(t)=\partial\etA^l_{i_l}/\partial\etA^{l-1}_{i_{l-1}}$ evaluated at $\alpha_t$, the contribution of a path $p=(i_0,\dots,i_L)$ is
\begin{equation}
\etC_p=\sum_{t=1}^{N_\alpha} w_t\,(\evx_{i_0}-\evx'_{i_0})\prod_{l=1}^{L}J^l_{i_l,i_{l-1}}(t).
\end{equation}
If we are taking the average contribution over a dataset, we can simply treat the batch dimension as more integration steps (perhaps with uneven weights), and the same algorithm applies.

We build a tree that partitions the path set. Each node is a prefix $(i_0,\dots,i_l)$, which stands for the set of all paths that start with $(i_0,\dots,i_l)$. The root is empty, which stands for all paths, and the leaves are individual paths. When we traverse the tree, at each node we compute a product for each integration step (its state):
\begin{equation}
v_t=w_t\,(\evx_{i_0}-\evx'_{i_0})\prod_{l'=1}^{l}J^{l'}_{i_{l'},i_{l'-1}}(t),\qquad t=1,\dots,N_\alpha.
\end{equation}
Each \emph{suffix} $(i_{l+1},\dots,i_L)$ also has a partial product $s_t=\prod_{l'>l}J^{l'}_{i_{l'},i_{l'-1}}(t)$, and when we extend a prefix by a suffix to obtain a full path, we get $\etC_p=\sum_t v_t s_t$. Therefore, at a prefix, we can bound the contribution of any path that extends it if we have a bound on $s_t$ for every step $t$.

\paragraph{The bound.} An important consideration in this algorithm is to come up with a tight bound. If the bound is loose, we will waste time expanding prefixes that do not lead to any of the top-$k$ paths. Because each integration step has its own product, a single bound over all steps is too loose, since there may only be one step where $s_t$ is large yet the bound thinks every step could be just as large. We therefore compute a bound for each step separately. For each step $t$, we precompute once, backward from the output, the largest magnitude $R^l_i(t)$ of $s_t$ over all suffixes that start at neuron $(l,i)$:
\begin{equation}
R^l_i(t)=\max_{j\in\gV_{l+1}}\big|J^{l+1}_{j,i}(t)\big|\,R^{l+1}_j(t),
\end{equation}
starting from $R^L=1$ at the output. With this bound on $|s_t|$, we have a bound on the contribution of any path that extends a prefix at $(l,i_l)$,
\begin{equation}
|\etC_p|\le\sum_{t=1}^{N_\alpha}|v_t|\,R^l_{i_l}(t),
\label{eq:prefix-bound}
\end{equation}
so we key the prefix by this value in the priority queue.

\paragraph{Expansion.} We organize the prefixes in a priority queue, ordered by the bound in \Eqref{eq:prefix-bound}. The algorithm pops the entry with the largest bound, expands it into its children with their own state $v_t\,J^{l+1}_{c,i_l}(t)$ and key, and pushes them back into the queue. A child that completes a path is pushed with the path's exact contribution $|\etC_p|=|\sum_t v_t|$ as its key in place of a bound. When a complete path is popped, it is returned.

\paragraph{Correctness.} Expanding a prefix partitions its paths among its children, so each path is covered by exactly one entry in the queue. Therefore, the search never misses a path and each path can only be returned once. Furthermore, the bound never grows along the tree: a child's bound $\sum_t|v_t|\,|J^{l+1}_{c,i_l}(t)|\,R^{l+1}_c(t)$ is at most its parent's $\sum_t|v_t|\,R^l_{i_l}(t)$, and a complete path's exact contribution is at most the bound of the prefix it came from. Therefore, when a complete path is popped from the queue, it is guaranteed to be stronger than any path not yet seen, since its exact contribution outranks every bound still in the queue. Since top paths get popped in order, we can stop the algorithm at any time and the paths already returned will be the top paths, making this an \emph{anytime} algorithm.

\paragraph{Pseudocode.}
\begin{enumerate}
\item Compute $R^l(t)$ in a backward pass from the output for every step $t$.
\item Push the root into the priority queue, whose children are the input neurons, with state $v_t=w_t(\evx-\evx')$.
\item Repeat until the queue is empty: Pop the entry with the largest key.
    \begin{enumerate}
    \item If it is a complete path, output it. If $k$ paths have been output, stop.
    \item Otherwise, expand it: for every child $c$, update $v_t\leftarrow v_t\,J_{c,\cdot}(t)$ and push the child with its bound, or, if it is a complete path, with its exact $|\etC_p|=|\sum_t v_t|$.
    \end{enumerate}
\end{enumerate}

\begin{figure}[ht]
\begin{center}
\includegraphics[width=\textwidth]{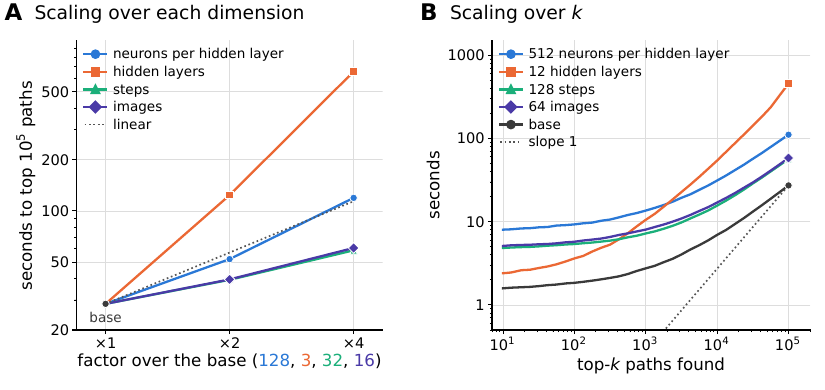}
\end{center}
\caption{Scaling of the top-$k$ algorithm on MNIST-trained MLPs, on an M3 Pro MacBook Pro with 36~GB of memory. The base run has 3 hidden layers of 128 neurons each, 32 integration steps, and the mean over 16 images. (A) Scaling over each of the four dimensions, with the others held at the base. (B) Scaling over $k$, the number of top paths to return, for the base run and each dimension maxed out.}
\label{fig:topk-scaling}
\end{figure}

\paragraph{Scaling.} Figure~\ref{fig:topk-scaling} measures the time to find the top $10^5$ paths through single neurons on MNIST-trained ReLU MLPs as each dimension of the problem is scaled up from a base run. Even though the algorithm does not have a worst-case guarantee, it is efficient in practice: it finds the top $10^5$ of the $784\times128^{12}\approx1.5\times10^{28}$ paths of a 12-layer MLP with 128 neurons per layer in 11 minutes on an M3 Pro MacBook Pro. In panel A, the time grows sublinearly in steps and images and about linearly in width. Over depth it grows superlinearly but remains reasonable, and tests at larger scales are needed to find where the algorithm stops being feasible. A caveat for the depth scaling: for the 12-layer run, the peak memory reached 80 GB on a 36 GB machine, which triggered macOS's compression and paging, which could have inflated the runtime. In panel B, we see that the time to find the top $k$ paths starts sublinear in $k$ and approaches linear as $k$ approaches $10^5$, sooner for deeper networks. An exhaustive algorithm needs to enumerate all paths, which is intractable for even modestly sized networks. The small discrepancy between matched settings in panels A and B is due to A including the time to free the queue after the search ends.

\paragraph{Limitations.} The algorithm we presented finds the top-$k$ paths in reasonable time on the scales we tested. However, scaling behavior on very deep and very wide nets remains untested due to resource constraints. For very deep nets, since the algorithm does not have an asymptotic worst-case guarantee (to our knowledge no such algorithm exists for DAG searches with sum-of-product path lengths), we do not know how tight the bound stays for more layers in practice. For very wide nets, memory starts to be prohibitive. For example, a pairwise Jacobian for GPT-2 small costs 2.5 TB in float32. Therefore, scaling to these models at the single-neuron resolution requires matrix-free Jacobians evaluated on the fly, which trades off time for space. In addition, one may want the top-$k$ paths not at the single neuron level, but at coarser granularity. In this case, each node contains a set of neurons, each having different bounds and interactions to adjacent layers, which makes bounding more challenging.

Even though still in experimental phase, our codebase provides some strategies to address these challenges. To save memory, we can push only the most promising child, together with one entry, the sibling entry, that stands for the remaining children under a shared bound, so the queue grows by about two entries per expansion instead of layer width. A sibling entry's bound can in turn be computed over blocks of integration steps and refined only when the entry reaches the top of the queue, which saves storing a bound per child and step. A signed interval on $s_t$ in place of its magnitude also tightens the bound on coarser granularities, when a unit groups several neurons, such as attention head-level or convolutional channel-level paths. On the nine runs of Figure~\ref{fig:topk-scaling}, with single neurons as units, the signed interval equals the magnitude bound, and sibling entries with refinement made the search 1.8 to 12 times slower while holding a much smaller queue (roughly 200 K entries at its peak in the base run, against 3.7 M). Even though our algorithm does not currently scale to the largest models and granularity cases under reasonable resources, we hope that the strategies we provide pave the way to more advanced (perhaps more specialized) algorithms that can scale to these cases in the future.

\section{Contribution Concentrates in a Sparse Set of \texorpdfstring{Top-$k$}{Top-k} Paths}
\label{sec:top-k-concentration}

\begin{figure}[ht]
    \centering
    \includegraphics[width=\textwidth]{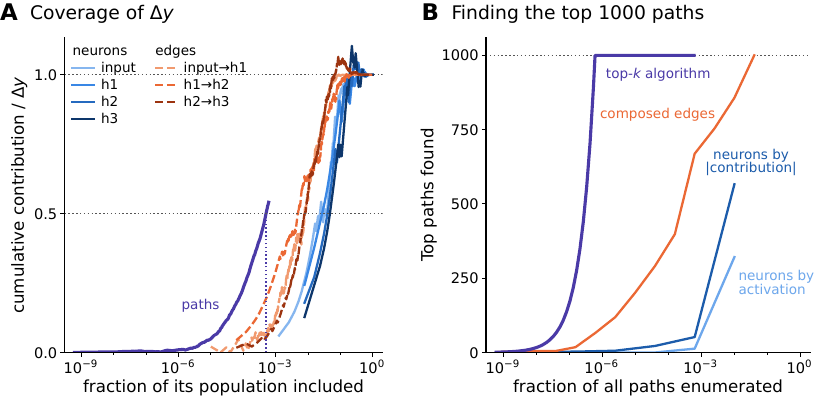}
    \caption{(A) Coverage of $\Delta y$ as a function of the fraction of a population included: top paths among all paths, top neurons among all neurons in a layer, and top edges among all edges in a layer. About 10 times fewer paths are needed by fraction to cover half of the output compared to edges, and another 10 times fewer than neurons. (B) The number of top paths found using a sequential beam search by most important neurons or edges. A beam search over edges needs to enumerate about 100 times more paths than requested, and neurons 100 times even more.}
    \label{fig:topk-statistics}
\end{figure}

Because of completeness, the output change $\Delta y$ equals the sum over all path contributions, and also the sum of neurons in a layer or edges in a layer. To interpret a network, it is often desirable to find a small set of objects that capture important properties of the whole network. Ultimately, the property to explain is the output, so a natural question is how much of $\Delta y$ concentrates in the top units among the population that distributes that sum. In \Figref{fig:topk-statistics}, we study the \emph{coverage} of a set of units, defined as their total contribution as a fraction of $\Delta y$, as a function of the number of top units included. For example, for neurons of a given layer, how many top contributing neurons do you need to explain the output difference? We measure coverage as a function of top units included, for the populations of paths, neurons, and edges. We find that in a 3-layer MLP with over a billion paths, only less than 0.1\% of top paths are needed to explain half of the output. Around 1\% of edges and close to 10\% of neurons in a layer are needed for the same coverage, suggesting that network computation is significantly sparser at the path level compared to the edge and neuron levels. In panel B, we address the question of whether a simple beam search over the most important neurons or edges in a layer sequentially yields the top paths. We find that beam search over the top edges needs to enumerate 100 times more paths than the number requested, and beam search over the top neurons 100 times even more to cover all the paths requested. Therefore, an efficient algorithm to find the top-$k$ paths directly is important for analyzing path contributions at scale.

\section{Prevalence of E-I Cancellation}
\label{sec:cancellation-prevalence}

\Figref{fig:silent-neuron} shows cancellation between pathways for one neuron. Here we ask how common cancellation is over all neurons of the same layer, over all input classes and output logits, and over networks trained from different random seeds.

We call a pair of an input class and an output logit an \emph{input-output scenario}, written $c$s@$k$: the validation images of class $c$, with contributions taken to logit $k$. There are $10\times 10=100$ scenarios, and 3s@3 is the one shown in \Figref{fig:silent-neuron}. For each scenario and each of the 128 neurons of h1, we divide the class-mean contribution of the neuron into total E and I effects at two levels: its 128 edges to the neurons of h2, and its 16,384 paths through h2 and h3. At each level, $E$ is the sum of the positive parts and $I$ the sum of the negative parts, so that $I\le 0$. The \emph{gross} contribution is $E-I=|E|+|I|$ and the \emph{net} contribution is $E+I$. By completeness, the net contribution at both levels is equal to the contribution of the neuron. We call the ratio of gross to absolute net contribution the \emph{cancellation}. It is 1 when all parts have the same sign, and it is large when a large E and a large I leave a small net contribution.

\begin{figure}[!htbp]
    \centering
    \includegraphics[width=\textwidth]{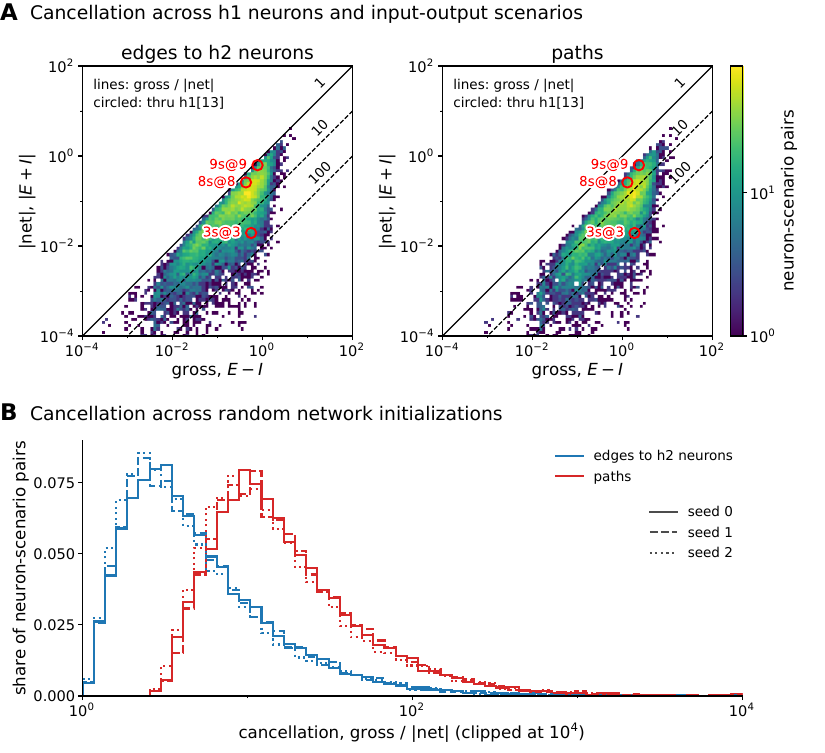}
    \caption{Cancellation between pathways across neurons, input-output scenarios and random seeds, in the setting of \Figref{fig:silent-neuron}. A scenario $c$s@$k$ is the validation images of class $c$ with contributions taken to logit $k$. For each h1 neuron in each of the 100 scenarios, the class-mean contribution is divided into the neuron's 128 edges to h2 neurons, or into its 16,384 paths through h2 and h3. $E$ and $I$ are the sums of the positive and negative contributions; the gross contribution is $E-I$ and the net contribution is $E+I$. (A) Two-dimensional histograms of the gross contribution against the absolute net contribution over all neuron-scenario pairs, for the network of \Figref{fig:silent-neuron}. Diagonal lines mark a ratio of gross to absolute net contribution of 1, 10 and 100. Red circles mark h1[13] in three scenarios of \Figref{fig:silent-neuron}E. (B) Distribution over neuron-scenario pairs of the ratio of gross to absolute net contribution, for networks trained from three random seeds.}
    \label{fig:mlp-cancellation}
\end{figure}

\Figref{fig:mlp-cancellation}A shows gross against net contribution for the network of \Figref{fig:silent-neuron}, over the 12,400 pairs of a neuron and a scenario in which the neuron has a nonzero gross contribution. Three scenarios of h1[13] shown in \Figref{fig:silent-neuron}E are circled. From edges to paths, the whole distribution moves to a larger gross contribution at the same net contribution, resulting in a cancellation factor roughly three times as large. Furthermore, we observe that across a horizontal line in each panel, there is a spread of gross contributions for the same net effect, showing that both our initial hypotheses---that the same net effect could be the result of the accumulation of small individual paths or the residual from strong but cancelling paths---are observed across neurons and scenarios. \Figref{fig:mlp-cancellation}B shows the distribution of the cancellation factor for three networks trained in the same way from different random seeds. The three distributions nearly coincide and show the consistent increase in cancellation from edges to paths, demonstrating that the phenomenon is not specific to a particular network or random seed.

\section{Comparison with Influence Paths, CODEC and Path Patching}
\label{sec:path-patching}

\paragraph{Conceptual comparison.} As stated with its definition, path contribution is mathematically equivalent to influence paths~\citep{luInfluencePathsCharacterizing2020}. That work and its extension to patterns~\citep{luInfluencePatternsExplaining2021} use the quantity to rank paths and to describe the route that information takes through a language model. However, they do not use the tool to measure cancellation effects at different granularities, and therefore our discovery of prevalent cancellation patterns across granularities is novel. Furthermore, they do not state or derive hierarchical completeness, which is a key property of Path Contribution that makes cancellation meaningful. Because the parts at any level (the neurons of a layer, the edges of a neuron, or its paths) sum to the same output difference, the total E, the total I and their cancellation are measured on one scale at every level and can be compared between levels, as in \Figref{fig:silent-neuron}E and \Figref{fig:mlp-cancellation}. CODEC~\citep{melanderCausalInterpretationNeural2026} quantifies the contributions of neurons and discovers motifs on the neuron level, and so fundamentally does not address path-level questions such as cancellation between paths that lead to a small net neuron contribution (\Figref{fig:silent-neuron}). Summing all the paths through a neuron recovers CODEC, so CODEC can be seen as a special case of Path Contribution at the neuron granularity.

Path patching~\citep{goldowsky-dillLocalizingModelBehavior2023,wangInterpretabilityWildCircuit2022} measures the causal effect of a set of paths by simulating an intervention: the paths in the set receive a counterfactual input, all other paths keep the original input, and the change in the output is recorded. Methods built on Integrated Gradients are also inherently causal measures. They ask the question: if the input starts from a baseline, chosen to represent a null expectation of the input, and changes in steps towards a target, which network components caused the output to change to the target output? The intervention at each integration step is the incremental change experienced by each unit or path, and the counterfactual is that component retaining its value from the previous step. The gradient of the output with respect to the component measures the causal effect of this intervention. A component’s contribution is the sum of these stepwise causal effects along the line from baseline to target, and the contributions of all components sum exactly to the total output change produced by moving the input from baseline to target. Thus, unlike patching, which measures one large counterfactual that can push the network outside its operating range, Integrated Gradients accumulates small counterfactuals along the path from baseline to target.

Path Contribution is thus an instance of these methods, measuring the causal effect of the most granular component, paths, which then recovers the causal effect of coarser components via partial summation. The chained gradient $\frac{\partial \etA^L_{i_L}}{\partial \etA^{L-1}_{i_{L-1}}} \cdots \frac{\partial \etA^1_{i_1}}{\partial \etA^0_{i_0}} \frac{\partial \etA^0_{i_0}}{\partial \alpha}$ is the infinitesimal causal effect of perturbing the input along the path $(i_0, i_1, \dots, i_L)$, where $i_0$ changes due to $\alpha$, $i_1$ changes due to $i_0$, and so on, independently from other paths because of the chain rule expansion. Path Contribution then accumulates these infinitesimal causal effects along the straight line from baseline to target to obtain a cumulative causal effect that enjoys the completeness property.

To compare path patching with path contribution experimentally, we take the counterfactual input to be the baseline, so that both methods explain the same output difference. We define the patching effect of a path as the output with the path on the original input minus the output with the path patched. This effect depends on which other paths are patched at the same time, a set that we call the \emph{context}. The \emph{remove-one} effect is the effect when no other path is patched, so that the rest of the network sees the original input. The \emph{keep-one} effect is the effect in the context where every other path is patched.

Importantly, none of these previous methods attempts to address the exponential number of paths. \citet{luInfluencePatternsExplaining2021} uses a beam search heuristic to find a sparse circuit, and not a guaranteed top-$k$ algorithm. We show in \Figref{fig:topk-statistics} that such a search enumerates $10^4$ times more paths than requested to find the true top paths in the 3-layer MLP. To our knowledge, the only other work that devises an exact search is \citet{duGeneralizingBackpropagationGradientBased2023}, but their path values are raw gradients with no integration, and they only find the top one path. Finding the top $k$ paths has enabled us to study their interactions at scale, and we showed that they reconstruct key properties of the network much more sparsely than neurons or edges: the output difference in \Figref{fig:topk-statistics}A and contribution pattern over scenarios \Figref{fig:silent-neuron}D,G.

\paragraph{The toy model.} In the toy model of \Figref{fig:toy}, giving each path its own copy of the input turns the network into $a_1+b_1-a_2b_2$, where $a_1$ and $b_1$ feed the direct paths and $a_2$ and $b_2$ the paths through $c$. Table~\ref{tab:toy-patching} lists the contribution and the patching effects of each path with a zero baseline. A direct path has the same patching effect in every context, equal to its contribution. A path through the product has an effect of $-ab$ when the other path is present and of $0$ when it is patched, and its contribution, $-ab/2$, is the average of the two. Remove-one effects count the product twice and keep-one effects never count it, so neither sums to the output; the contributions do by completeness. Here the contribution of a path equals its Shapley value~\citep{rozemberczkiShapleyValueMachine2022}, the average of its patching effect over all contexts, which also enjoys completeness but in general requires iterating over $2^{k-1}$ contexts for $k$ paths. Path Contribution and Shapley values are equal whenever the network is multilinear in the paths' input copies~\citep{owenMultilinearExtensionsGames1972}. On linear networks remove-one path patching, Path Contribution, and Shapley values all agree, but on the toy model only Path Contribution and Shapley values do. In general, there is no guarantee that the three agree, and we can view Path Contribution as a representative statistic of the distribution of a path's causal effect over contexts that guarantees completeness and is efficient to compute compared to Shapley values.

\begin{table}[!htbp]
\caption{Path contribution and patching effects for the four paths of the toy model $d=a+b-a\times b$, with a zero baseline. Remove one is the difference between the full model and just that path patched, and keep one is the difference between just that path and the empty model.}
\label{tab:toy-patching}
\centering
\begin{tabular}{lcccc}
\toprule
Path & Contribution & Remove one & Keep one & Shapley value \\
\midrule
$a\rightarrow d$ & $a$ & $a$ & $a$ & $a$ \\
$a\rightarrow c\rightarrow d$ & $-ab/2$ & $-ab$ & $0$ & $-ab/2$ \\
$b\rightarrow c\rightarrow d$ & $-ab/2$ & $-ab$ & $0$ & $-ab/2$ \\
$b\rightarrow d$ & $b$ & $b$ & $b$ & $b$ \\
\midrule
Sum & $a+b-ab$ & $a+b-2ab$ & $a+b$ & $a+b-ab$ \\
\bottomrule
\end{tabular}
\end{table}

\begin{figure}[!htbp]
    \centering
    \includegraphics[width=\textwidth]{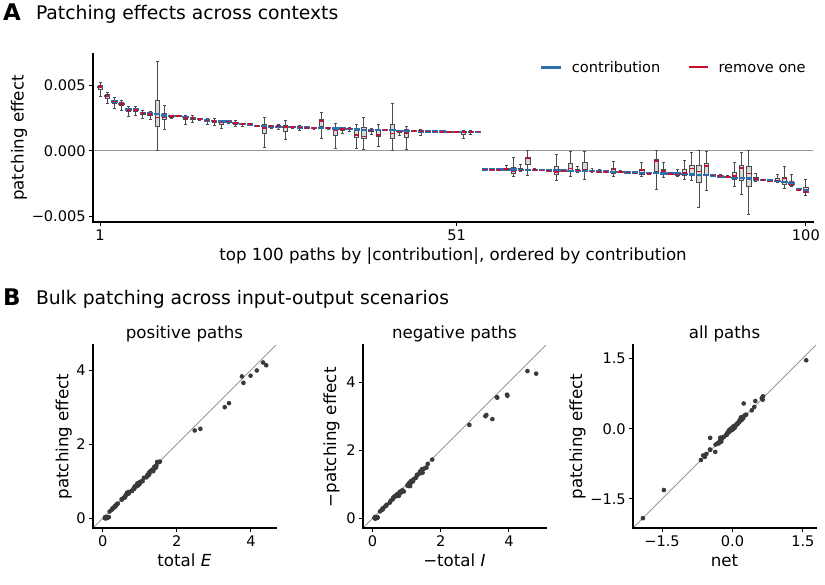}
    \caption{Path contribution against path patching effect, for the paths of h1[13] in the network of \Figref{fig:silent-neuron}, with the black baseline as the counterfactual input. The patching effect is the output keeping the paths in question minus the output with the paths patched, averaged over images in the scenario. (A) The 100 paths with the largest absolute class-mean contribution in 3s@3, ordered by contribution. Gray boxes show the distribution of the patching effect of each path over 3,001 contexts: each of the other 127 neurons of h1 is set to its baseline value with probability 0.1, 0.25 or 0.5 (1,000 contexts each), plus the context with no other neuron patched, which gives the remove-one effect (red). Boxes are quartiles and whiskers extend to 1.5 times the interquartile range; outliers are not drawn. Blue: the path contribution. (B) Each point is one of the 100 input-output scenarios. Left: the effect of patching all paths with a positive class-mean contribution, against the total E. Middle: minus the effect of patching all paths with a negative contribution, against minus the total I. Right: the effect of patching all 16,384 paths, against the net contribution. The gray line is the diagonal.}
    \label{fig:path-patching}
\end{figure}

\paragraph{Patching effects across contexts.} \Figref{fig:path-patching}A compares the path patching and Path Contribution in the setting of \Figref{fig:silent-neuron}. We take the 100 paths of h1[13] with the largest absolute class-mean contribution and measure the patching effect of each, averaged over all images of 3s across contexts. A context sets each of the other 127 neurons of h1 to its value at the baseline, independently with probability $q$, which patches every path through that neuron. We see that the contribution follows the patching effect closely, and the difference between them is much smaller than many variations across contexts. The two methods thus measure closely very similar scores in practice. The contribution acts as one summary over contexts and, unlike the patching effect in any one context, it satisfies completeness.

\paragraph{Bulk patching.} We further show that the effect of patching many paths at once agrees with the total contribution of those paths. For each of the 100 input-output scenarios of Appendix~\ref{sec:cancellation-prevalence}, we patch on every image all paths of h1[13] whose class-mean contribution is positive, all paths whose contribution is negative, and all 16,384 paths. \Figref{fig:path-patching}B shows that the three patching effects fall closely on the diagonal against the total E, the total I and the net contribution, demonstrating that total contribution is a good predictor of the effect of bulk patching.

\section{Robustness and Generalizability of the Retinal Findings}
\label{sec:retina-robustness}
The retina results in the main text describe one ganglion cell, one baseline and one trained model. Here we test whether they depend on those choices: we repeat the analysis for all $36$ recorded cells, check the numerical integration, change the baseline, retrain the model from a different random seed, and follow the decomposition over training. Unless stated otherwise, cells are evaluated on a natural-scene block shown ten times and used for neither training nor validation.

\begin{figure}[!htbp]
    \centering
    \includegraphics[width=\textwidth]{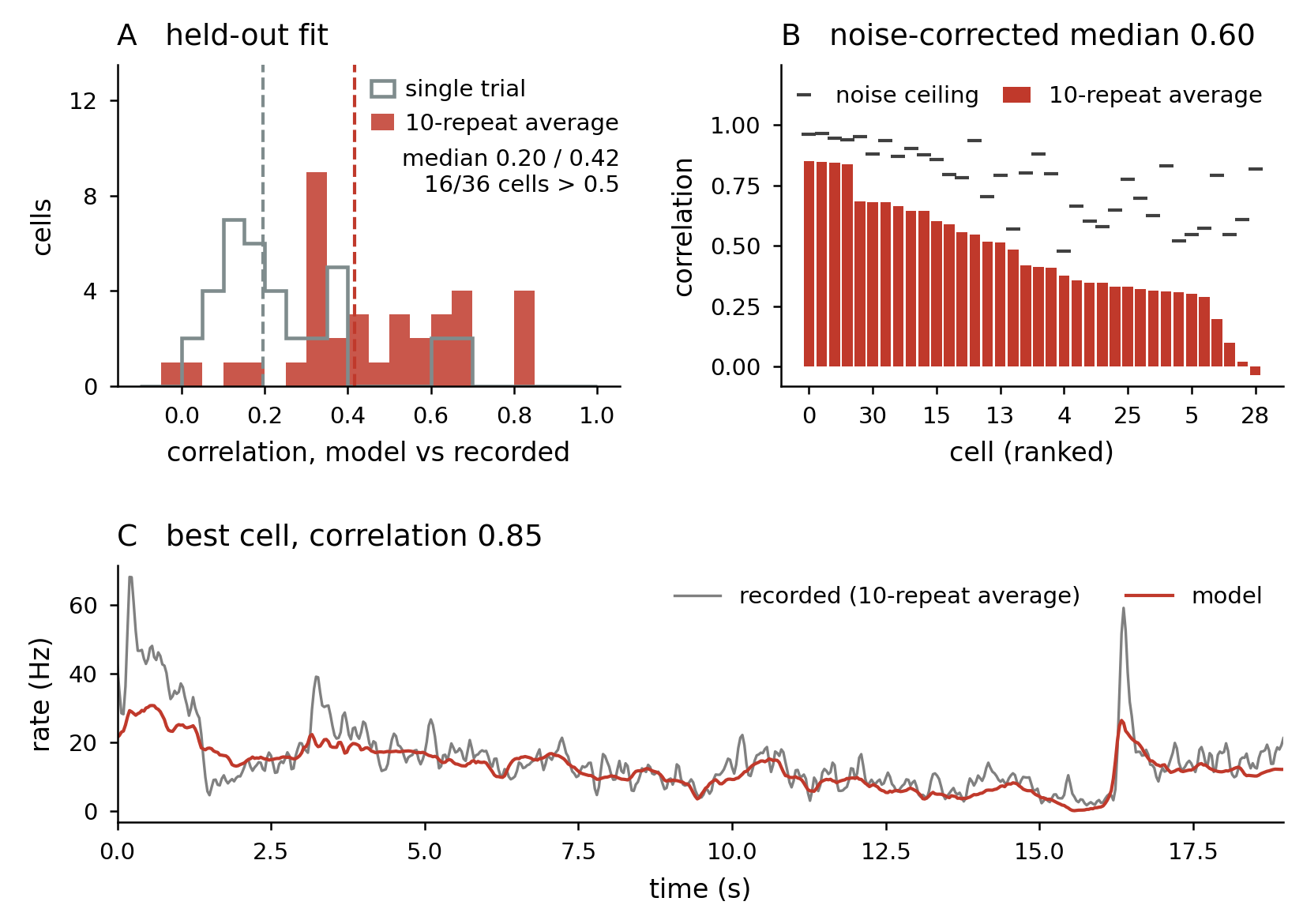}
    \caption{Fit quality for all $36$ cells on the held-out repeat block. (A) Correlation between the predicted rate and the recorded rate, for a single trial and for the average over ten repeats; dashed lines are the medians. (B) Correlation of each cell with the ten-repeat average, with the cell's noise ceiling, the square root of the split-half reliability of the average. (C) Predicted and recorded rate of the best-predicted cell.}
    \label{fig:retina-fit}
\end{figure}

\paragraph{Fit quality.} Across the $36$ cells, the predicted rate correlates with the response averaged over the ten repeats at a median of $0.42$ and a maximum of $0.85$, and $16$ cells exceed $0.5$ (\Figref{fig:retina-fit}). Part of the remaining gap is noise in the recording rather than error in the model: correcting for the split-half reliability (the correlation between two halves of a trial set to measure internal consistency) of the ten-repeat average raises the median to $0.60$ and the maximum to $0.89$.

\paragraph{Choice of baseline.} A baseline is a counterfactual input: path contribution explains the difference between the output on the stimulus and the output on the baseline, $\sum_{c_0,c_1} P_t[c_0,c_1] = y_t - y_{\mathrm{base}}$. The natural counterfactual for a movie is its mean image, each pixel averaged over the clip and repeated over the $30$ frames of the model's input; this is the default baseline used for the retinal models. Replacing it with a uniform gray screen at the mean luminance leaves the decomposition unchanged (\Figref{fig:retina-baseline}). Integrated IRFs agree at a median cosine similarity of $0.99$, first-PC loadings at $0.98$, and the ranking of the $256$ paths correlates at $0.96$. For the main-text cell, the first two PCs carry $79.5\%$ and $79.9\%$ of the variance, and the first circuit reproduces the IRF at a cosine of $0.96$ under both baselines.
Cancellation is present under every baseline we tried, with a median factor of $6.2$ under the mean image, $8.2$ under gray and $6.4$ under a uniform black screen.

\begin{figure}[!htbp]
    \centering
    \includegraphics[width=\textwidth]{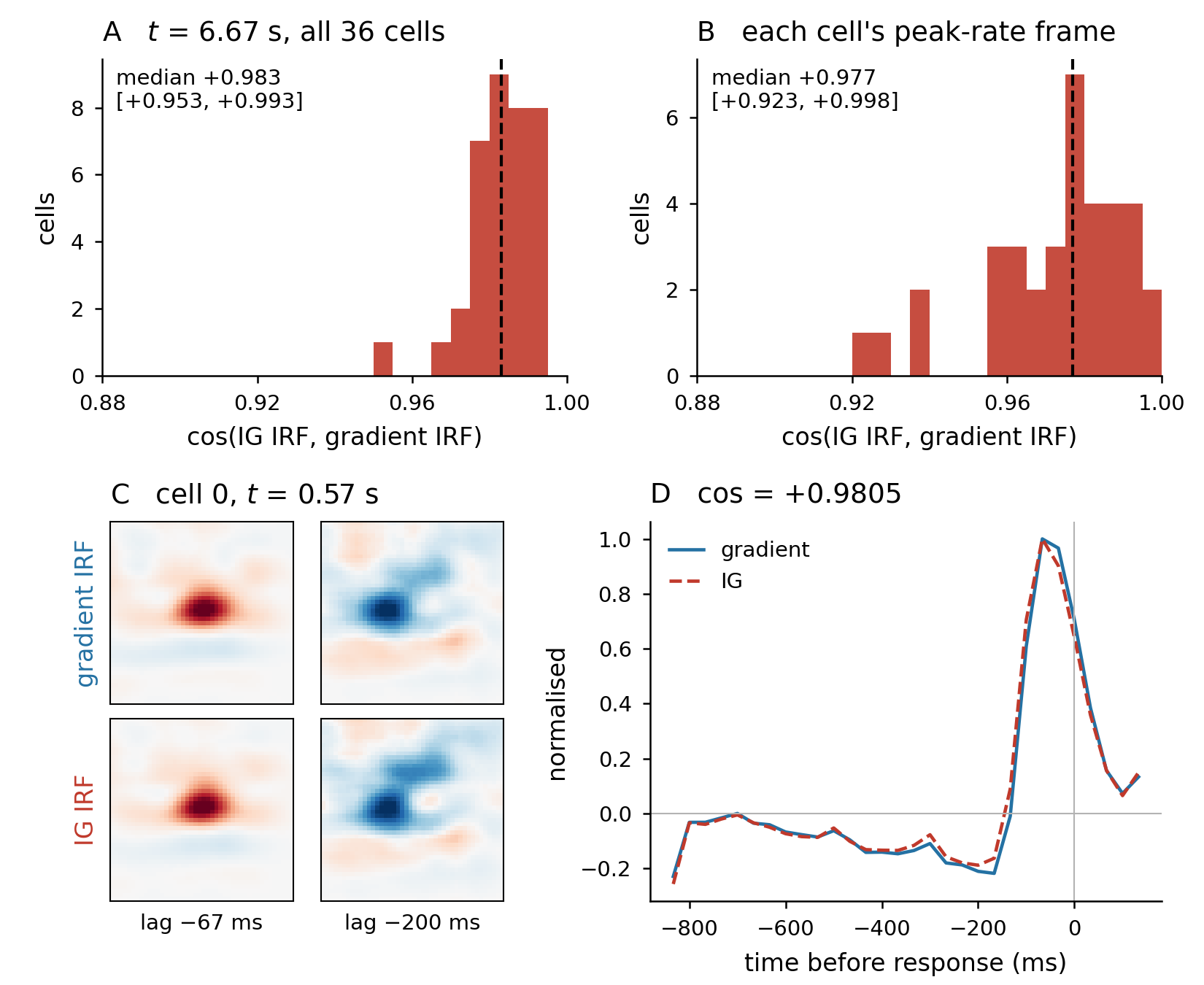}
    \caption{The IRF from integrated path contribution matches the gradient IRF. (A) Cosine similarity between the two fields for all $36$ cells at one frame. (B) The same at each cell's frame of peak firing. (C) Both fields of the best-predicted cell at its frame of peak firing, at two lags before the response. (D) Time course of both fields, each projected on its own spatial pattern at the strongest lag.}
    \label{fig:retina-ig}
\end{figure}

\paragraph{Integrated and plain receptive fields.} The instantaneous receptive field (IRF) of a cell is the gradient of its rate with respect to the stimulus \citep{maheswaranathanInterpretingRetinalNeural2023}. Path contribution integrates this gradient along the line from the baseline to the stimulus, so the field it implies is the gradient accumulated over that line. The two RFs
agree in every cell, with a median cosine similarity of $0.98$ both at a fixed frame and at each cell's frame of peak firing (\Figref{fig:retina-ig}). The integration therefore leaves the receptive field it decomposes unchanged.

\begin{figure}[!htbp]
    \centering
    \includegraphics[width=\textwidth]{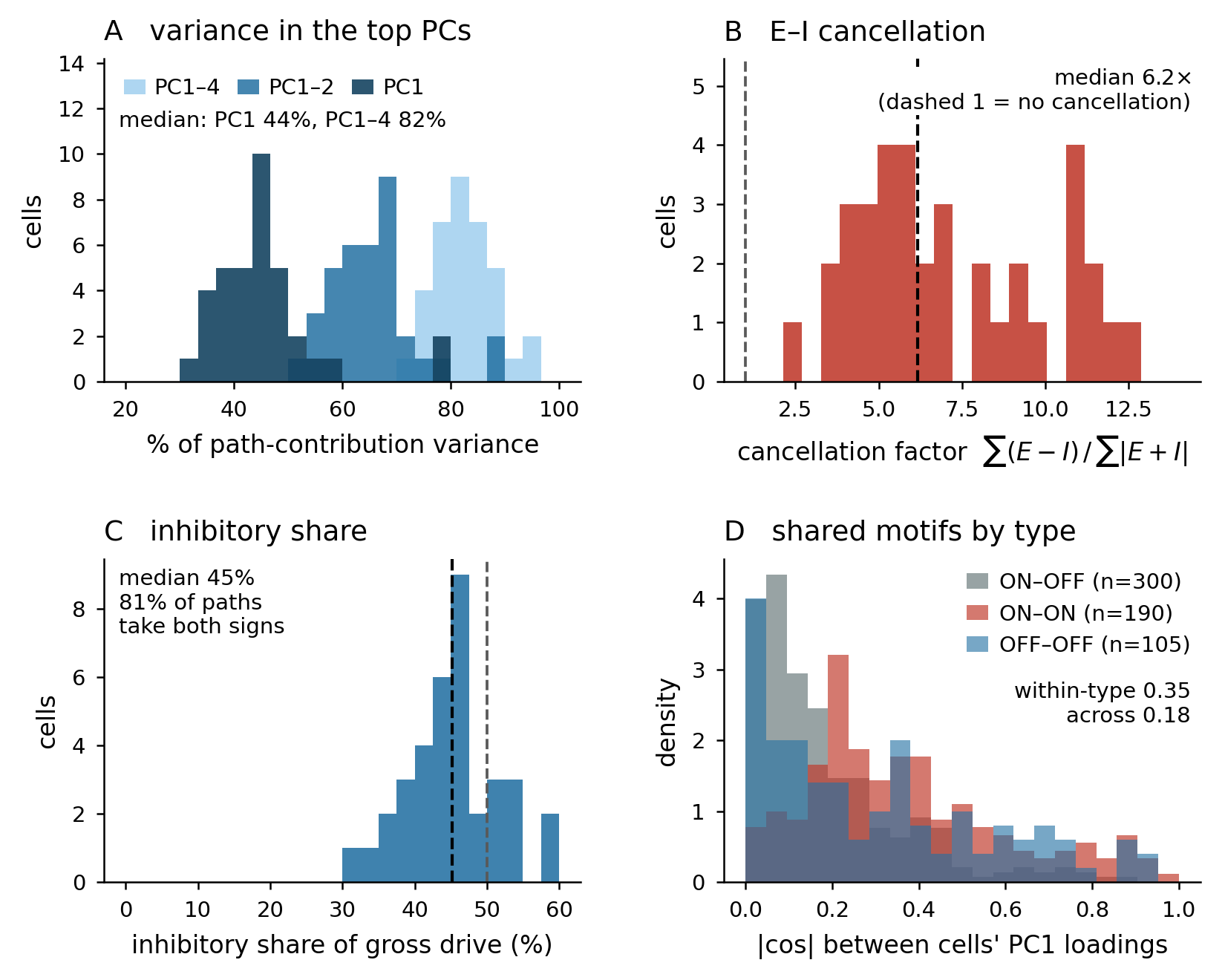}
    \caption{Low-dimensional circuits and E--I cancellation in recorded ganglion cells. (A) Variance of the path contributions carried by the first PC, the first two and the first four. (B) E--I cancellation factor, $\sum_t (E_t - I_t) / \sum_t |E_t + I_t|$, the gross drive per unit of net output; $1$ means no cancellation. (C) Share of the gross drive that is inhibitory. (D) Absolute cosine similarity between the first-PC loadings of pairs of cells, for ON--ON, OFF--OFF and ON--OFF pairs.}
    \label{fig:retina-population}
\end{figure}

\paragraph{Generalizability across cells.} The two features of the decomposition in the main text hold in every cell (\Figref{fig:retina-population}). First, a few circuits dominate: the first PC of the $256$ path contributions carries a median
of $44\%$ of their variance, and the first four carry $82\%$. Second, the paths cancel: their gross drive exceeds the net output by a median factor of $6.2$ (range $2.7$--$12.8$), and inhibition supplies $45\%$ of it. The factor varies from cell to cell, as expected if different cells perform different computations.

\begin{figure}[!htbp]
    \centering
    \includegraphics[width=\textwidth]{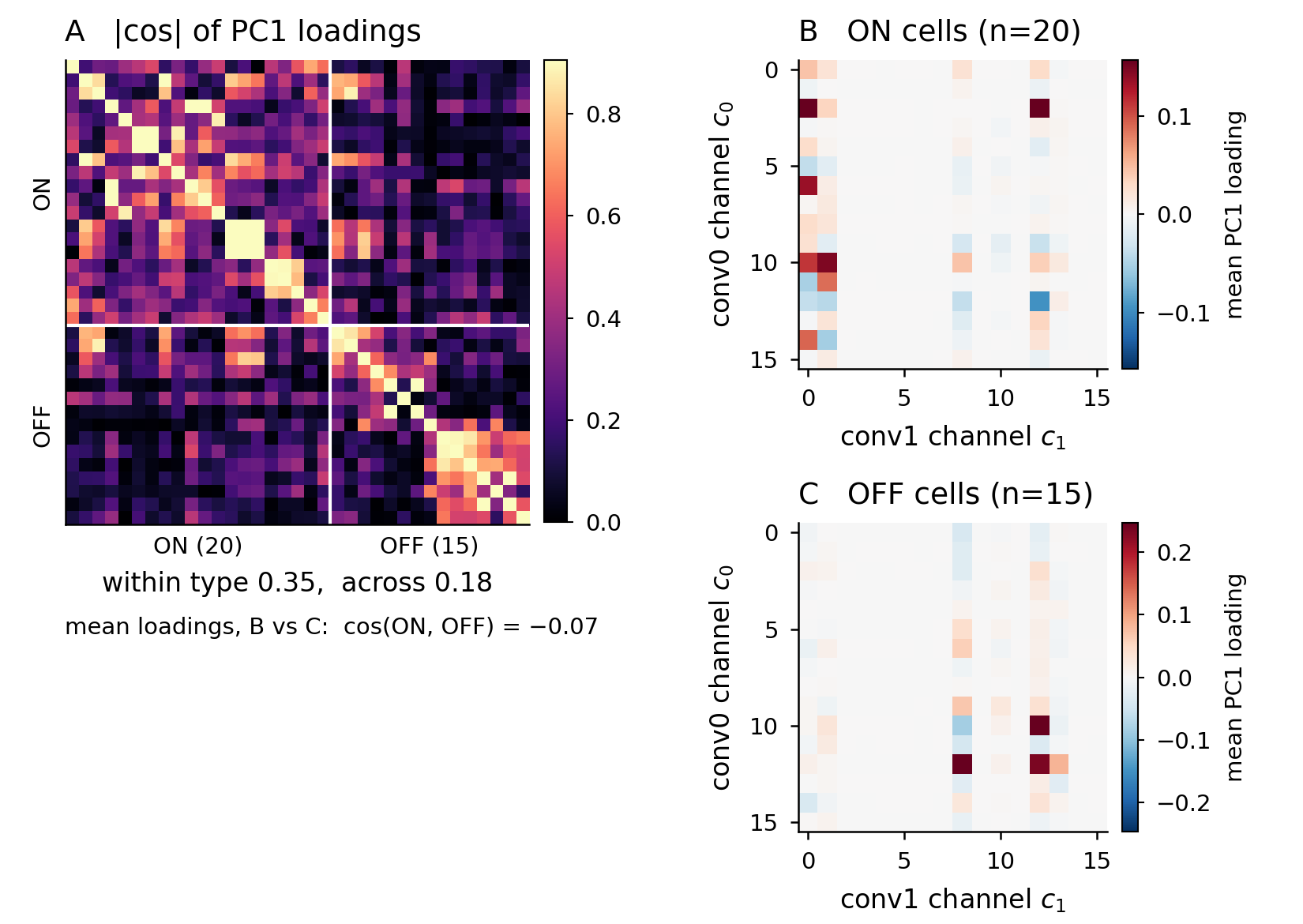}
    \caption{Path motifs are shared within, not across, cell polarity, for the $35$ cells whose white-noise and natural-scene receptive fields agree in polarity. (A) Absolute cosine similarity between the first-PC loadings of every pair of cells, ON cells first. (B) Mean first-PC loading of the ON cells over the $16 \times 16$ paths. (C) The same for the OFF cells.}
    \label{fig:retina-onoff}
\end{figure}

\begin{figure}[!htbp]
    \centering
    \includegraphics[width=\textwidth]{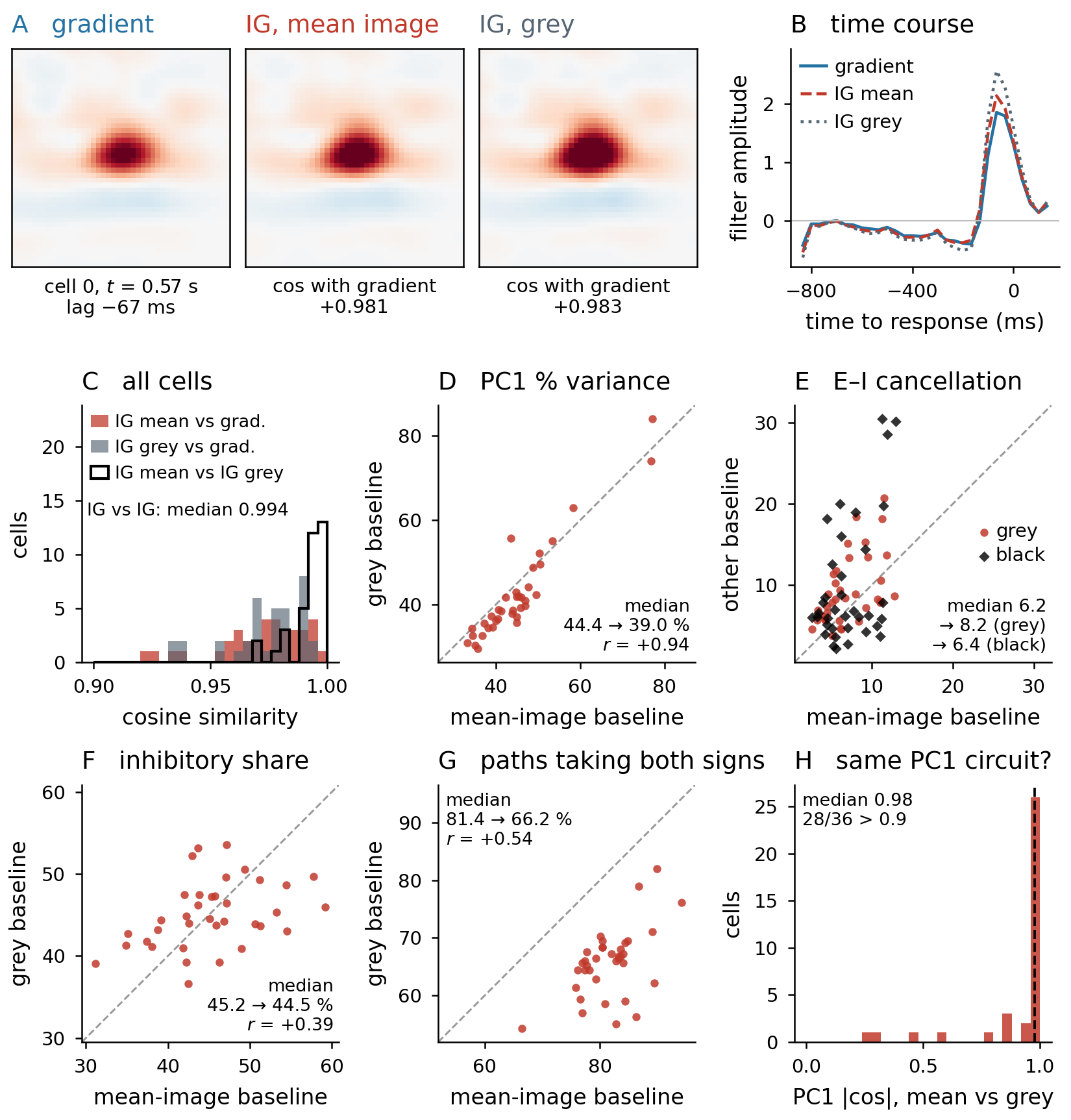}
    \caption{Mean-image baseline against a uniform gray baseline. (A) Gradient IRF and integrated-gradients IRF under each baseline, for the best-predicted cell at its frame of peak firing, on one color scale. (B) Time course of the three fields. (C) Cosine similarity between the fields over all $36$ cells. (D--G) Per-cell first-PC variance, E--I cancellation factor, inhibitory share and fraction of paths taking both signs under the two baselines; points on the diagonal are unchanged. (H) Absolute cosine similarity between each cell's first-PC loadings under the two baselines.}
    \label{fig:retina-baseline}
\end{figure}

\paragraph{ON and OFF cells.} If the circuits recovered by path contribution reflect the computation of a cell, cells of the same type should have similar circuits and be distinct from other complementary cell types. In the retina, an elemental classification of ganglion cells is between ON ganglion cells (fire when light increases) and OFF ganglion cells (fire when light decreases). We took cells where the polarities agreed between the white-noise receptive field and the natural scene IRF ($20$ ON with positive polarity, $15$ OFF with negative polarity). We found that cells of the same type share a dominant circuit: the mean absolute cosine similarity between first-PC loadings is higher within type ($0.35$) compared to across types ($0.18$, $p<0.00025$, permutation test shuffling the ON/OFF labels $4,000$ times), and the mean ON and OFF circuits are nearly orthogonal (cosine $-0.07$) (\Figref{fig:retina-population}D, \Figref{fig:retina-onoff}). This illustrates that path contribution can help find circuits that are selective for complementary network computations in the retina.

\begin{figure}[!htbp]
    \centering
    \includegraphics[width=0.8\textwidth]{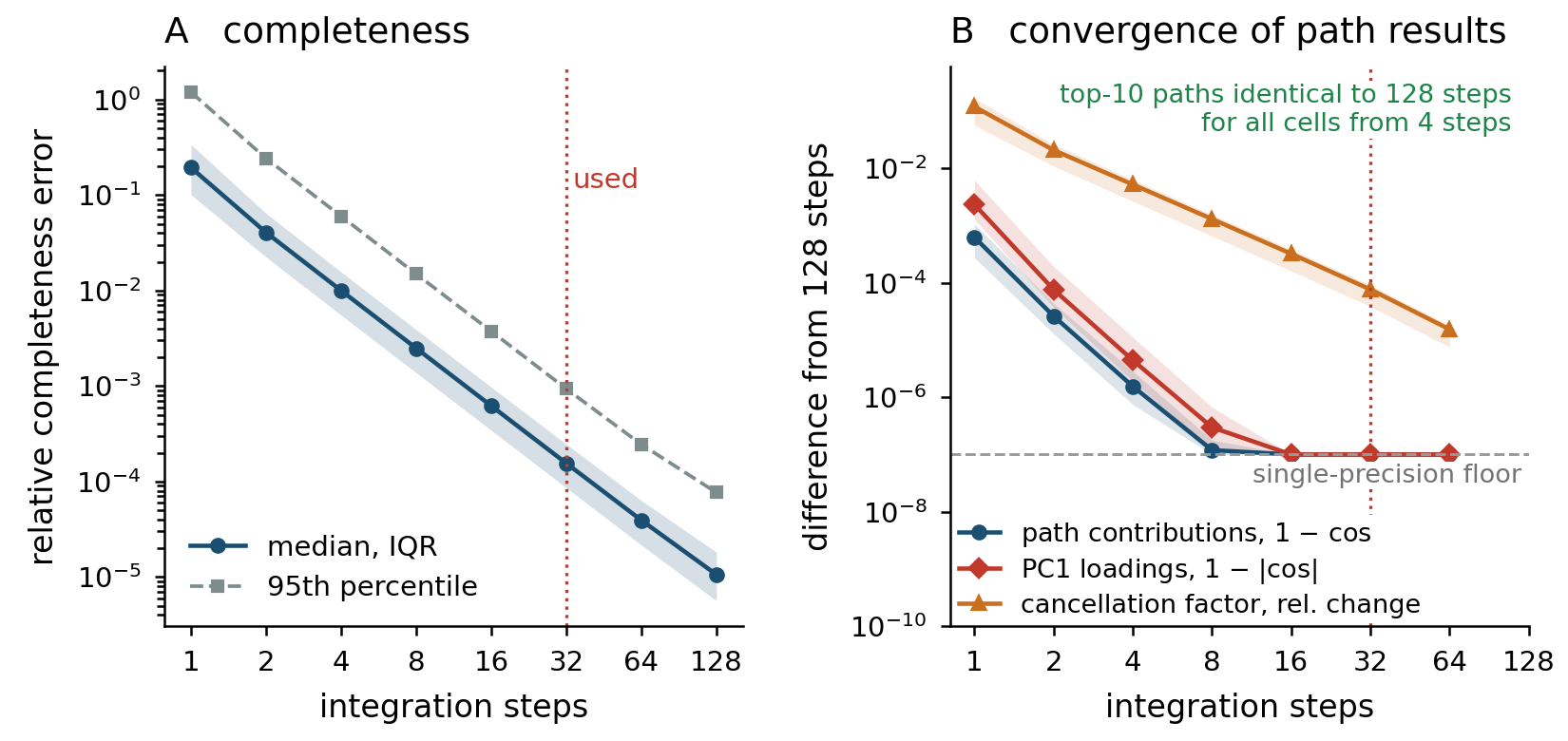}
    \caption{Integration steps, for all $36$ cells on every sixth window of the held-out block; lines are medians and shading the interquartile range. (A) Relative completeness error against the number of trapezoid steps. (B) Difference from the $128$-step result of the contributions at each frame, the first-PC loadings and the cancellation factor; the dashed line marks single-precision accuracy. Dotted lines mark the $32$ steps used throughout.}

    \label{fig:retina-steps}
\end{figure}

\paragraph{Integration steps.} Path contributions are integrated gradients over the input, computed with the trapezoid rule over a number of integration steps. It is unknown whether the path contributions are sensitive and variable with respect to the number of integration steps, especially in the context of the retinal model. To measure the integration error, we recomputed the contributions of all $36$ cells with  fewer integration steps from the same test data (\Figref{fig:retina-steps}). At $32$ steps the contributions sum to $y_t - y_{\mathrm{base}}$ with a median relative error of $1.6\times10^{-4}$, and the error falls fourfold with every doubling of the steps. The path results converge much sooner: the ten strongest paths of every cell are identical from $4$ steps to $128$, and between $32$ and $128$ steps the cancellation factor changes by $7\times10^{-5}$. The numerical error is far below any effect we report.

\begin{figure}[!htbp]
    \centering
    \includegraphics[width=\textwidth]{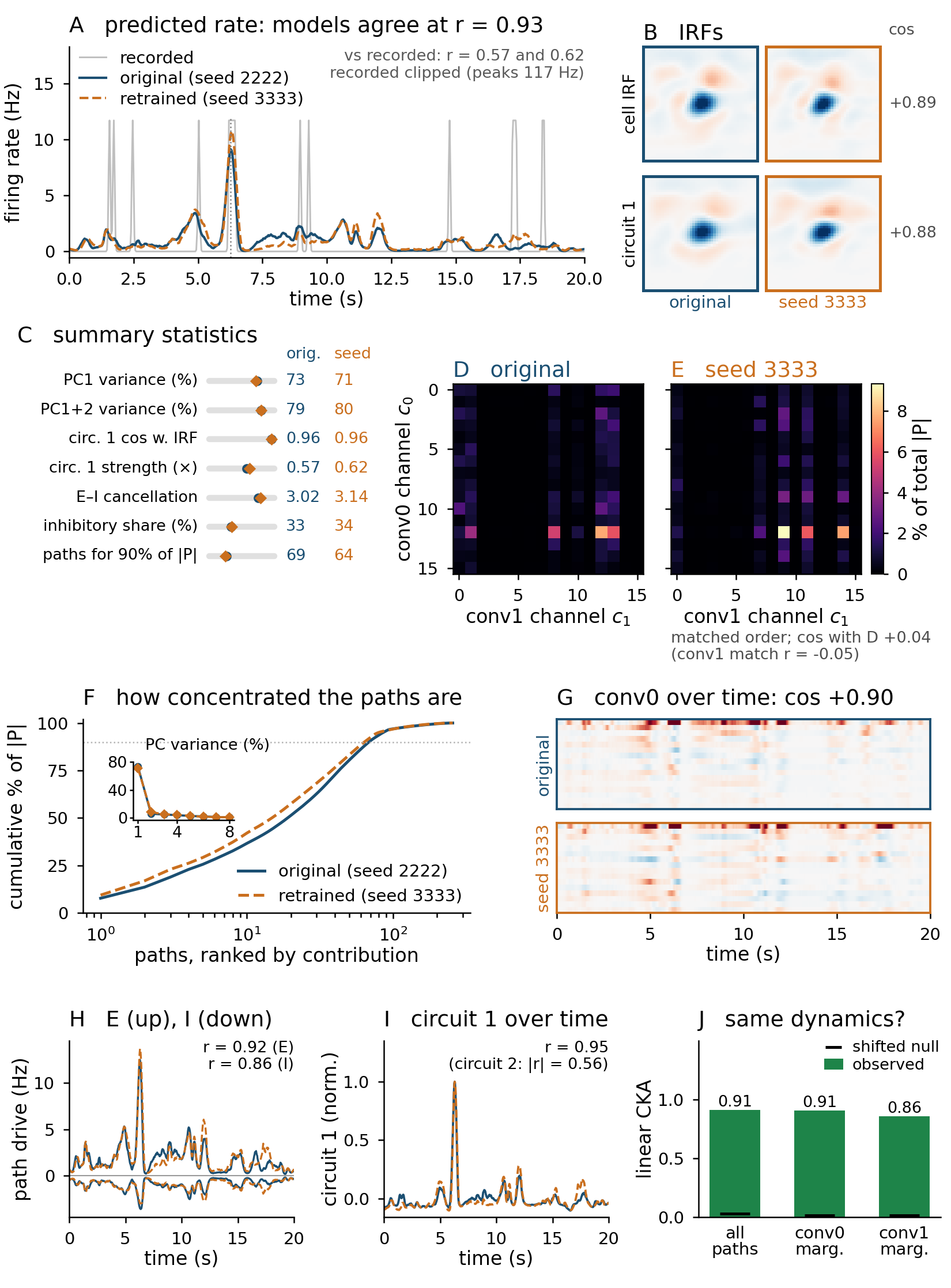}
    \caption{Two models trained on the same cells and data from different random seeds, for the cell and stimulus of the main text. (A) Recorded rate and the rate predicted by each model; the recorded trace is clipped. (B) Each model's IRF and first-circuit IRF at the frame of the main text. (C) Summary statistics of the path decomposition. (D) Mean absolute contribution of each path in the original model. (E) The same in the second model, with its channels reordered to their matched counterparts in the original. (F) Cumulative contribution of the paths ranked by size, and variance of the leading PCs. (G) First-layer contributions over time, matched channels in the same order. (H) Summed excitatory (upward) and inhibitory (downward) drive. (I) Activation of the first circuit over time. (J) Linear centered kernel alignment between the two models' path contributions, against a time-shifted control.}
    \label{fig:retina-seed}
\end{figure}

\paragraph{Random seed.} A second model trained on the same cells and data from a different random seed is evaluated to  see whether results are comparable to the retinal model discussed in the main text. Its validation correlation matches the original's (median $0.52$ against $0.51$), and for the main-text cell the predicted rates of the two models correlate at $0.93$ (\Figref{fig:retina-seed}). It also yields the same decomposition: the first two PCs carry $79.5\%$ and $79.7\%$ of the variance, the first circuit reproduces the IRF at a cosine of $0.96$ in both models, and the cancellation factor is $3.0$ and $3.1$. The labels of hidden channels are arbitrary, since nothing in training fixes them, so we compare the path dynamics with linear CKA \citep{kornblith2019similarity}, which is invariant to any relabeling or rotation of the paths. It is $0.91$ between the two models, against $0.03$ for a time-shifted control. Path contribution therefore recovers the same circuits and dynamics from independently trained models.

\begin{figure}[!htbp]
    \centering
    \includegraphics[width=\textwidth]{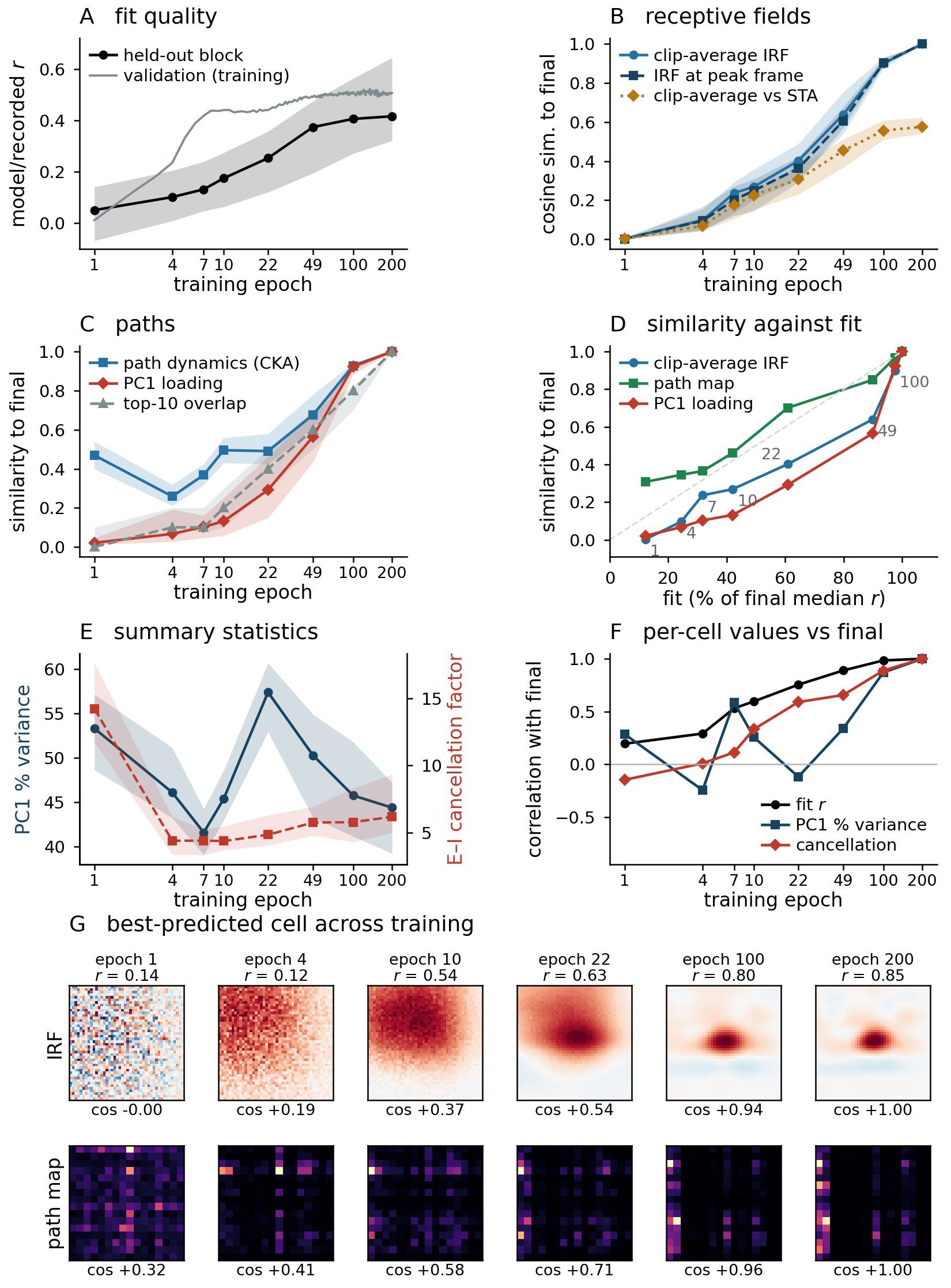}
    \caption{The same network over training, for all $36$ cells on the held-out block; lines are medians over cells, shading the interquartile range. (A) Correlation with the recorded response, on the held-out block and on the validation block during training. (B) Cosine similarity to the final model of the clip-average and peak-frame IRFs, and of the clip-average IRF to the white-noise STA. (C) Similarity of the paths to the final model: the linear CKA over time, $|\cos|$ of the first-PC loadings, and top-10 path overlap. (D) Similarity to the final model against fit (\% of final median correlation); numbers are epochs. (E) First-PC variance and cancellation factor. (F) Correlation across cells with the final values. (G) IRF and path map of the best-predicted cell, with its correlation and each map's cosine similarity to the final model.}
    \label{fig:retina-training}
\end{figure}

\paragraph{Fit quality and the path results.} To ask how fit quality affects the path results with the cells, stimulus and network held fixed, we recomputed the receptive fields and paths of the same network at seven checkpoints during
training (\Figref{fig:retina-training}). We measure fit as the median correlation between predicted and recorded rates on the held-out repeat block. Both converge with the fit, and the paths converge faster than the receptive field. At epoch $22$, the median correlation is $0.25$, or $61\%$ of the final model's $0.42$. A cell's path map already has a cosine similarity of $0.70$ with its final map, while its IRF reaches $0.40$. By epoch $100$, the median correlation is $0.41$, or $98\%$ of the final value, and the path maps, first-PC loadings and IRFs have cosine similarities of $0.97$, $0.92$ and $0.90$ with their final form, and each cell's cancellation factor correlates with its final value at $0.89$. The path results are therefore stable once the model fits the data.

\section{Additional Figures}

\begin{figure}[ht]
    \centering
    \includegraphics[width=\textwidth]{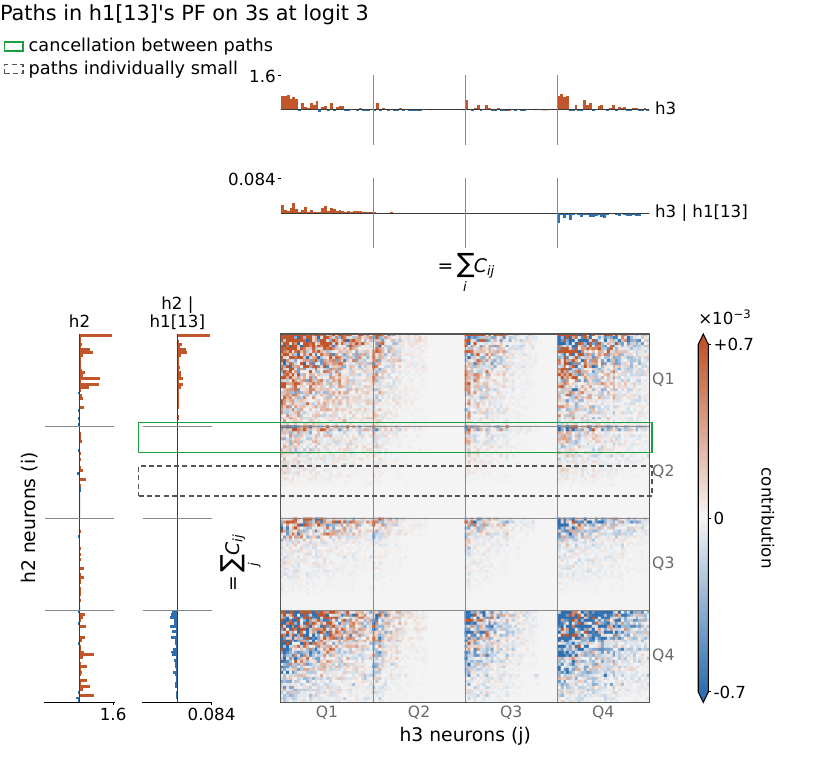}
    \caption{A larger version of \Figref{fig:silent-neuron}F.}
    \label{fig:silent-neuron-matrix}
\end{figure}

\end{document}

%% file: math_commands.tex
\usepackage{amsmath,amsfonts,bm}

\def\Figref#1{Figure~\ref{#1}}

\def\eqref#1{equation~\ref{#1}}
\def\Eqref#1{Equation~\ref{#1}}

\def\1{\bm{1}}

\def\ve{{\bm{e}}}

\def\vm{{\bm{m}}}

\def\vx{{\bm{x}}}

\def\evm{{m}}

\def\evx{{x}}

\def\mM{{\bm{M}}}

\DeclareMathAlphabet{\mathsfit}{\encodingdefault}{\sfdefault}{m}{sl}
\SetMathAlphabet{\mathsfit}{bold}{\encodingdefault}{\sfdefault}{bx}{n}
\newcommand{\tens}[1]{\bm{\mathsfit{#1}}}
\def\tA{{\tens{A}}}

\def\gE{{\mathcal{E}}}

\def\gG{{\mathcal{G}}}

\def\gV{{\mathcal{V}}}

\def\sS{{\mathbb{S}}}

\def\emM{{M}}

\newcommand{\etens}[1]{\mathsfit{#1}}

\def\etA{{\etens{A}}}

\def\etC{{\etens{C}}}



%% file: references.bib
@article{elhage2021mathematical,
   title={A Mathematical Framework for Transformer Circuits},
   author={Elhage, Nelson and Nanda, Neel and Olsson, Catherine and Henighan, Tom and Joseph, Nicholas and Mann, Ben and Askell, Amanda and Bai, Yuntao and Chen, Anna and Conerly, Tom and DasSarma, Nova and Drain, Dawn and Ganguli, Deep and Hatfield-Dodds, Zac and Hernandez, Danny and Jones, Andy and Kernion, Jackson and Lovitt, Liane and Ndousse, Kamal and Amodei, Dario and Brown, Tom and Clark, Jack and Kaplan, Jared and McCandlish, Sam and Olah, Chris},
   year={2021},
   journal={Transformer Circuits Thread},
   note={https://transformer-circuits.pub/2021/framework/index.html}
}

@inproceedings{dosovitskiyImageWorth16x162020,
  title = {An {{Image}} Is {{Worth}} 16x16 {{Words}}: {{Transformers}} for {{Image Recognition}} at {{Scale}}},
  shorttitle = {An {{Image}} Is {{Worth}} 16x16 {{Words}}},
  booktitle = {International {{Conference}} on {{Learning Representations}}},
  author = {Dosovitskiy, Alexey and Beyer, Lucas and Kolesnikov, Alexander and Weissenborn, Dirk and Zhai, Xiaohua and Unterthiner, Thomas and Dehghani, Mostafa and Minderer, Matthias and Heigold, Georg and Gelly, Sylvain and Uszkoreit, Jakob and Houlsby, Neil},
  year = 2020,
  month = oct,
  urldate = {2025-11-18},
  langid = {english}
}

@inproceedings{mengLocatingEditingFactual2022,
  title = {Locating and {{Editing Factual Associations}} in {{GPT}}},
  booktitle = {Advances in {{Neural Information Processing Systems}}},
  author = {Meng, Kevin and Bau, David and Andonian, Alex J. and Belinkov, Yonatan},
  year = 2022,
  month = oct,
  urldate = {2026-09-24},
  langid = {english}
}

@inproceedings{anconaBetterUnderstandingGradientbased2018,
  title = {Towards Better Understanding of Gradient-Based Attribution Methods for {{Deep Neural Networks}}},
  booktitle = {International {{Conference}} on {{Learning Representations}}},
  author = {Ancona, Marco and Ceolini, Enea and {\"O}ztireli, Cengiz and Gross, Markus},
  year = 2018,
  month = feb,
  urldate = {2026-08-17},
  langid = {english}
}

@article{baccusRetinalCircuitThat2008,
  title = {A {{Retinal Circuit That Computes Object Motion}}},
  author = {Baccus, Stephen A. and {\"O}lveczky, Bence P. and Manu, Mihai and Meister, Markus},
  year = 2008,
  month = jul,
  journal = {Journal of Neuroscience},
  volume = {28},
  number = {27},
  pages = {6807--6817},
  publisher = {Society for Neuroscience},
  issn = {0270-6474, 1529-2401},
  doi = {10.1523/JNEUROSCI.4206-07.2008},
  urldate = {2026-04-18},
  chapter = {Articles},
  copyright = {Copyright \copyright{} 2008 Society for Neuroscience 0270-6474/08/286807-11\$15.00/0},
  langid = {english},
  pmid = {18596156}
}

@article{bachPixelWiseExplanationsNonLinear2015,
  title = {On {{Pixel-Wise Explanations}} for {{Non-Linear Classifier Decisions}} by {{Layer-Wise Relevance Propagation}}},
  author = {Bach, Sebastian and Binder, Alexander and Montavon, Gr{\'e}goire and Klauschen, Frederick and M{\"u}ller, Klaus-Robert and Samek, Wojciech},
  year = 2015,
  month = jul,
  journal = {PLOS ONE},
  volume = {10},
  number = {7},
  pages = {e0130140},
  publisher = {Public Library of Science},
  issn = {1932-6203},
  doi = {10.1371/journal.pone.0130140},
  urldate = {2026-09-24},
  langid = {english}
}

@inproceedings{conmyAutomatedCircuitDiscovery2023a,
  title = {Towards {{Automated Circuit Discovery}} for {{Mechanistic Interpretability}}},
  booktitle = {Thirty-Seventh {{Conference}} on {{Neural Information Processing Systems}}},
  author = {Conmy, Arthur and {Mavor-Parker}, Augustine N. and Lynch, Aengus and Heimersheim, Stefan and {Garriga-Alonso}, Adri{\`a}},
  year = 2023,
  month = nov,
  urldate = {2026-05-08},
  langid = {english}
}

@article{deboerTriggeredCorrelation1968,
  title = {Triggered {{Correlation}}},
  author = {De Boer, Egbert and Kuyper, Paul},
  year = 1968,
  month = jul,
  journal = {IEEE Transactions on Biomedical Engineering},
  volume = {BME-15},
  number = {3},
  pages = {169--179},
  issn = {1558-2531},
  doi = {10.1109/TBME.1968.4502561},
  urldate = {2026-09-24}
}

@inproceedings{dhamdhereHowImportantNeuron2018,
  title = {How {{Important}} Is a {{Neuron}}},
  booktitle = {International {{Conference}} on {{Learning Representations}}},
  author = {Dhamdhere, Kedar and Sundararajan, Mukund and Yan, Qiqi},
  year = 2018,
  month = sep,
  urldate = {2026-09-24},
  langid = {english}
}

@inproceedings{duGeneralizingBackpropagationGradientBased2023,
  title = {Generalizing {{Backpropagation}} for {{Gradient-Based Interpretability}}},
  booktitle = {Proceedings of the 61st {{Annual Meeting}} of the {{Association}} for {{Computational Linguistics}} ({{Volume}} 1: {{Long Papers}})},
  author = {Du, Kevin and Torroba Hennigen, Lucas and Stoehr, Niklas and Warstadt, Alex and Cotterell, Ryan},
  editor = {Rogers, Anna and {Boyd-Graber}, Jordan and Okazaki, Naoaki},
  year = 2023,
  month = jul,
  pages = {11979--11995},
  publisher = {Association for Computational Linguistics},
  address = {Toronto, Canada},
  doi = {10.18653/v1/2023.acl-long.669},
  urldate = {2026-08-17}
}

@misc{geigerCausalAbstractionsNeural2021,
  title = {Causal {{Abstractions}} of {{Neural Networks}}},
  author = {Geiger, Atticus and Lu, Hanson and Icard, Thomas and Potts, Christopher},
  year = 2021,
  month = oct,
  number = {arXiv:2106.02997},
  eprint = {2106.02997},
  primaryclass = {cs.AI},
  publisher = {arXiv},
  doi = {10.48550/arXiv.2106.02997},
  urldate = {2026-07-02},
  archiveprefix = {arXiv}
}

@misc{goldowsky-dillLocalizingModelBehavior2023,
  title = {Localizing {{Model Behavior}} with {{Path Patching}}},
  author = {{Goldowsky-Dill}, Nicholas and MacLeod, Chris and Sato, Lucas and Arora, Aryaman},
  year = 2023,
  month = may,
  number = {arXiv:2304.05969},
  eprint = {2304.05969},
  primaryclass = {cs.LG},
  publisher = {arXiv},
  doi = {10.48550/arXiv.2304.05969},
  urldate = {2026-06-25},
  archiveprefix = {arXiv}
}

@misc{hatefiAttributionGuidedPruningInsight2026,
  title = {Attribution-{{Guided Pruning}} for {{Insight}} and {{Control}}: {{Circuit Discovery}} and {{Targeted Correction}} in {{Small}}-Scale {{LLMs}}},
  shorttitle = {Attribution-{{Guided Pruning}} for {{Insight}} and {{Control}}},
  author = {Hatefi, Sayed Mohammad Vakilzadeh and Dreyer, Maximilian and Achtibat, Reduan and Kahardipraja, Patrick and Wiegand, Thomas and Samek, Wojciech and Binder, Alexander and Lapuschkin, Sebastian},
  year = 2026,
  month = may,
  number = {arXiv:2506.13727},
  eprint = {2506.13727},
  primaryclass = {cs.LG},
  publisher = {arXiv},
  doi = {10.48550/arXiv.2506.13727},
  urldate = {2026-11-18},
  archiveprefix = {arXiv}
}

@misc{hatefiPruningExplainingRevisited2024,
  title = {Pruning {{By Explaining Revisited}}: {{Optimizing Attribution Methods}} to {{Prune CNNs}} and {{Transformers}}},
  shorttitle = {Pruning {{By Explaining Revisited}}},
  author = {Hatefi, Sayed Mohammad Vakilzadeh and Dreyer, Maximilian and Achtibat, Reduan and Wiegand, Thomas and Samek, Wojciech and Lapuschkin, Sebastian},
  year = 2024,
  month = oct,
  number = {arXiv:2408.12568},
  eprint = {2408.12568},
  primaryclass = {cs},
  publisher = {arXiv},
  doi = {10.48550/arXiv.2408.12568},
  urldate = {2025-11-20},
  archiveprefix = {arXiv}
}

@misc{kramarAtPEfficientScalable2024,
  title = {{{AtP}}*: {{An}} Efficient and Scalable Method for Localizing {{LLM}} Behaviour to Components},
  shorttitle = {{{AtP}}*},
  author = {Kram{\'a}r, J{\'a}nos and Lieberum, Tom and Shah, Rohin and Nanda, Neel},
  year = 2024,
  month = mar,
  number = {arXiv:2403.00745},
  eprint = {2403.00745},
  primaryclass = {cs},
  publisher = {arXiv},
  doi = {10.48550/arXiv.2403.00745},
  urldate = {2025-03-12},
  archiveprefix = {arXiv}
}

@article{lehkyNetworkModelShapefromshading1988a,
  title = {Network Model of Shape-from-Shading: Neural Function Arises from Both Receptive and Projective Fields},
  shorttitle = {Network Model of Shape-from-Shading},
  author = {Lehky, Sidney R. and Sejnowski, Terrence J.},
  year = 1988,
  month = jun,
  journal = {Nature},
  volume = {333},
  number = {6172},
  pages = {452--454},
  publisher = {Nature Publishing Group},
  issn = {1476-4687},
  doi = {10.1038/333452a0},
  urldate = {2026-09-24},
  copyright = {1988 Springer Nature Limited},
  langid = {english}
}

@inproceedings{luInfluencePathsCharacterizing2020,
  title = {Influence {{Paths}} for {{Characterizing Subject-Verb Number Agreement}} in {{LSTM Language Models}}},
  booktitle = {Proceedings of the 58th {{Annual Meeting}} of the {{Association}} for {{Computational Linguistics}}},
  author = {Lu, Kaiji and Mardziel, Piotr and Leino, Klas and Fredrikson, Matt and Datta, Anupam},
  editor = {Jurafsky, Dan and Chai, Joyce and Schluter, Natalie and Tetreault, Joel},
  year = 2020,
  month = jul,
  pages = {4748--4757},
  publisher = {Association for Computational Linguistics},
  address = {Online},
  doi = {10.18653/v1/2020.acl-main.430},
  urldate = {2026-08-17}
}

@inproceedings{luInfluencePatternsExplaining2021,
  title = {Influence {{Patterns}} for {{Explaining Information Flow}} in {{BERT}}},
  booktitle = {Advances in {{Neural Information Processing Systems}}},
  author = {Lu, Kaiji and Wang, Zifan and Mardziel, Piotr and Datta, Anupam},
  year = 2021,
  month = nov,
  urldate = {2026-09-19},
  langid = {english}
}

@inproceedings{lundbergUnifiedApproachInterpreting2017,
  title = {A {{Unified Approach}} to {{Interpreting Model Predictions}}},
  booktitle = {Advances in {{Neural Information Processing Systems}}},
  author = {Lundberg, Scott M and Lee, Su-In},
  year = 2017,
  volume = {30},
  publisher = {Curran Associates, Inc.},
  urldate = {2026-09-24}
}

@article{maheswaranathanInterpretingRetinalNeural2023,
  title = {Interpreting the Retinal Neural Code for Natural Scenes: {{From}} Computations to Neurons},
  shorttitle = {Interpreting the Retinal Neural Code for Natural Scenes},
  author = {Maheswaranathan, Niru and McIntosh, Lane T. and Tanaka, Hidenori and Grant, Satchel and Kastner, David B. and Melander, Joshua B. and Nayebi, Aran and Brezovec, Luke E. and Wang, Julia H. and Ganguli, Surya and Baccus, Stephen A.},
  year = 2023,
  month = sep,
  journal = {Neuron},
  volume = {111},
  number = {17},
  pages = {2742-2755.e4},
  issn = {0896-6273},
  doi = {10.1016/j.neuron.2023.06.007},
  urldate = {2024-10-25}
}

@inproceedings{marksSparseFeatureCircuits2024,
  title = {Sparse {{Feature Circuits}}: {{Discovering}} and {{Editing Interpretable Causal Graphs}} in {{Language Models}}},
  shorttitle = {Sparse {{Feature Circuits}}},
  booktitle = {The {{Thirteenth International Conference}} on {{Learning Representations}}},
  author = {Marks, Samuel and Rager, Can and Michaud, Eric J. and Belinkov, Yonatan and Bau, David and Mueller, Aaron},
  year = 2024,
  month = oct,
  urldate = {2026-08-17},
  langid = {english}
}

@inproceedings{melanderCausalInterpretationNeural2026,
  title = {Causal {{Interpretation}} of {{Neural Network Computations}} with {{Contribution Decomposition}}},
  booktitle = {The {{Fourteenth International Conference}} on {{Learning Representations}}, {{ICLR}} 2026, {{Rio}} de {{Janeiro}}, {{Brazil}}, {{April}} 23-27, 2026},
  author = {Melander, Joshua Brendan and Alaoui, Zaki and Liu, Shenghua and Ganguli, Surya and Baccus, Stephen},
  year = 2026,
  month = apr,
  urldate = {2026-02-11},
  langid = {english}
}

@article{olahZoomIntroductionCircuits2020,
  title = {Zoom {{In}}: {{An Introduction}} to {{Circuits}}},
  shorttitle = {Zoom {{In}}},
  author = {Olah, Chris and Cammarata, Nick and Schubert, Ludwig and Goh, Gabriel and Petrov, Michael and Carter, Shan},
  year = 2020,
  month = mar,
  journal = {Distill},
  volume = {5},
  number = {3},
  pages = {e00024.001},
  issn = {2476-0757},
  doi = {10.23915/distill.00024.001},
  urldate = {2025-09-25},
  langid = {english}
}

@article{olveczkySegregationObjectBackground2003,
  title = {Segregation of Object and Background Motion in the Retina},
  author = {{\"O}lveczky, Bence P. and Baccus, Stephen A. and Meister, Markus},
  year = 2003,
  month = may,
  journal = {Nature},
  volume = {423},
  number = {6938},
  pages = {401--408},
  publisher = {Nature Publishing Group},
  issn = {1476-4687},
  doi = {10.1038/nature01652},
  urldate = {2026-08-17},
  copyright = {2003 Macmillan Magazines Ltd.},
  langid = {english}
}

@misc{shrikumarComputationallyEfficientMeasures2018,
  title = {Computationally {{Efficient Measures}} of {{Internal Neuron Importance}}},
  author = {Shrikumar, Avanti and Su, Jocelin and Kundaje, Anshul},
  year = 2018,
  month = jul,
  number = {arXiv:1807.09946},
  eprint = {1807.09946},
  primaryclass = {cs.LG},
  publisher = {arXiv},
  doi = {10.48550/arXiv.1807.09946},
  urldate = {2026-09-24},
  archiveprefix = {arXiv}
}

@inproceedings{shrikumarLearningImportantFeatures2017,
  title = {Learning Important Features through Propagating Activation Differences},
  booktitle = {Proceedings of the 34th {{International Conference}} on {{Machine Learning}} - {{Volume}} 70},
  author = {Shrikumar, Avanti and Greenside, Peyton and Kundaje, Anshul},
  year = 2017,
  month = aug,
  series = {{{ICML}}'17},
  pages = {3145--3153},
  publisher = {JMLR.org},
  address = {Sydney, NSW, Australia},
  urldate = {2025-05-14}
}

@misc{simonyanDeepConvolutionalNetworks2014,
  title = {Deep {{Inside Convolutional Networks}}: {{Visualising Image Classification Models}} and {{Saliency Maps}}},
  shorttitle = {Deep {{Inside Convolutional Networks}}},
  author = {Simonyan, Karen and Vedaldi, Andrea and Zisserman, Andrew},
  year = 2014,
  month = apr,
  number = {arXiv:1312.6034},
  eprint = {1312.6034},
  primaryclass = {cs},
  publisher = {arXiv},
  doi = {10.48550/arXiv.1312.6034},
  urldate = {2025-05-14},
  archiveprefix = {arXiv}
}

@misc{smilkovSmoothGradRemovingNoise2017,
  title = {{{SmoothGrad}}: Removing Noise by Adding Noise},
  shorttitle = {{{SmoothGrad}}},
  author = {Smilkov, Daniel and Thorat, Nikhil and Kim, Been and Vi{\'e}gas, Fernanda and Wattenberg, Martin},
  year = 2017,
  month = jun,
  number = {arXiv:1706.03825},
  eprint = {1706.03825},
  primaryclass = {cs},
  publisher = {arXiv},
  doi = {10.48550/arXiv.1706.03825},
  urldate = {2025-03-10},
  archiveprefix = {arXiv}
}

@inproceedings{sundararajanAxiomaticAttributionDeep2017,
  title = {Axiomatic Attribution for Deep Networks},
  booktitle = {Proceedings of the 34th {{International Conference}} on {{Machine Learning}} - {{Volume}} 70},
  author = {Sundararajan, Mukund and Taly, Ankur and Yan, Qiqi},
  year = 2017,
  month = aug,
  series = {{{ICML}}'17},
  pages = {3319--3328},
  publisher = {JMLR.org},
  address = {Sydney, NSW, Australia},
  urldate = {2024-10-27}
}

@inproceedings{syedAttributionPatchingOutperforms2024,
  title = {Attribution {{Patching Outperforms Automated Circuit Discovery}}},
  booktitle = {Proceedings of the 7th {{BlackboxNLP Workshop}}: {{Analyzing}} and {{Interpreting Neural Networks}} for {{NLP}}},
  author = {Syed, Aaquib and Rager, Can and Conmy, Arthur},
  editor = {Belinkov, Yonatan and Kim, Najoung and Jumelet, Jaap and Mohebbi, Hosein and Mueller, Aaron and Chen, Hanjie},
  year = 2024,
  month = nov,
  pages = {407--416},
  publisher = {Association for Computational Linguistics},
  address = {Miami, Florida, US},
  doi = {10.18653/v1/2024.blackboxnlp-1.25},
  urldate = {2026-08-17}
}

@inproceedings{tanakaDeepLearningMechanistic2019a,
  title = {From Deep Learning to Mechanistic Understanding in Neuroscience: The Structure of Retinal Prediction},
  shorttitle = {From Deep Learning to Mechanistic Understanding in Neuroscience},
  booktitle = {Advances in {{Neural Information Processing Systems}}},
  author = {Tanaka, Hidenori and Nayebi, Aran and Maheswaranathan, Niru and McIntosh, Lane and Baccus, Stephen and Ganguli, Surya},
  year = 2019,
  volume = {32},
  publisher = {Curran Associates, Inc.},
  urldate = {2025-05-21}
}

@article{tziavelisAnykAlgorithmsEnumerating2025,
  title = {Any-k {{Algorithms}} for {{Enumerating Ranked Answers}} to {{Conjunctive Queries}}},
  author = {Tziavelis, Nikolaos and Gatterbauer, Wolfgang and Riedewald, Mirek},
  year = 2025,
  month = sep,
  journal = {ACM Transactions on Database Systems},
  volume = {51},
  number = {1},
  pages = {1:1--1:47},
  issn = {0362-5915},
  doi = {10.1145/3734517},
  urldate = {2026-07-14}
}

@inproceedings{vigInvestigatingGenderBias2020,
  title = {Investigating {{Gender Bias}} in {{Language Models Using Causal Mediation Analysis}}},
  booktitle = {Advances in {{Neural Information Processing Systems}}},
  author = {Vig, Jesse and Gehrmann, Sebastian and Belinkov, Yonatan and Qian, Sharon and Nevo, Daniel and Singer, Yaron and Shieber, Stuart},
  year = 2020,
  volume = {33},
  pages = {12388--12401},
  publisher = {Curran Associates, Inc.},
  urldate = {2026-09-04}
}

@article{walkerInceptionLoopsDiscover2019,
  title = {Inception Loops Discover What Excites Neurons Most Using Deep Predictive Models},
  author = {Walker, Edgar Y. and Sinz, Fabian H. and Cobos, Erick and Muhammad, Taliah and Froudarakis, Emmanouil and Fahey, Paul G. and Ecker, Alexander S. and Reimer, Jacob and Pitkow, Xaq and Tolias, Andreas S.},
  year = 2019,
  month = dec,
  journal = {Nature Neuroscience},
  volume = {22},
  number = {12},
  pages = {2060--2065},
  publisher = {Nature Publishing Group},
  issn = {1546-1726},
  doi = {10.1038/s41593-019-0517-x},
  urldate = {2026-09-24},
  copyright = {2019 The Author(s), under exclusive licence to Springer Nature America, Inc.},
  langid = {english}
}

@misc{wangInterpretabilityWildCircuit2022,
  title = {Interpretability in the {{Wild}}: A {{Circuit}} for {{Indirect Object Identification}} in {{GPT-2}} Small},
  shorttitle = {Interpretability in the {{Wild}}},
  author = {Wang, Kevin and Variengien, Alexandre and Conmy, Arthur and Shlegeris, Buck and Steinhardt, Jacob},
  year = 2022,
  month = nov,
  number = {arXiv:2211.00593},
  eprint = {2211.00593},
  primaryclass = {cs},
  publisher = {arXiv},
  doi = {10.48550/arXiv.2211.00593},
  urldate = {2025-09-25},
  archiveprefix = {arXiv}
}

@article{ameisen2025circuit,
  author={Ameisen, Emmanuel and Lindsey, Jack and Pearce, Adam and Gurnee, Wes and Turner, Nicholas L. and Chen, Brian and Citro, Craig and Abrahams, David and Carter, Shan and Hosmer, Basil and Marcus, Jonathan and Sklar, Michael and Templeton, Adly and Bricken, Trenton and McDougall, Callum and Cunningham, Hoagy and Henighan, Thomas and Jermyn, Adam and Jones, Andy and Persic, Andrew and Qi, Zhenyi and Ben Thompson, T. and Zimmerman, Sam and Rivoire, Kelley and Conerly, Thomas and Olah, Chris and Batson, Joshua},
  title={Circuit Tracing: Revealing Computational Graphs in Language Models},
  journal={Transformer Circuits Thread},
  year={2025},
  url={https://transformer-circuits.pub/2025/attribution-graphs/methods.html}
}

@article{bricken2023monosemanticity,
       title={Towards Monosemanticity: Decomposing Language Models With Dictionary Learning},
       author={Bricken, Trenton and Templeton, Adly and Batson, Joshua and Chen, Brian and Jermyn, Adam and Conerly, Tom and Turner, Nick and Anil, Cem and Denison, Carson and Askell, Amanda and Lasenby, Robert and Wu, Yifan and Kravec, Shauna and Schiefer, Nicholas and Maxwell, Tim and Joseph, Nicholas and Hatfield-Dodds, Zac and Tamkin, Alex and Nguyen, Karina and McLean, Brayden and Burke, Josiah E and Hume, Tristan and Carter, Shan and Henighan, Tom and Olah, Christopher},
       year={2023},
       journal={Transformer Circuits Thread},
       note={https://transformer-circuits.pub/2023/monosemantic-features/index.html}
    }

@article{ratliff1959responses,
  title={The responses of Limulus optic nerve fibers to patterns of illumination on the receptor mosaic},
  author={Ratliff, Floyd and Hartline, H Keffer},
  journal={The Journal of general physiology},
  volume={42},
  number={6},
  pages={1241--1255},
  year={1959},
  publisher={Rockefeller University Press}
}

@article{kuffler1953discharge,
  title={Discharge patterns and functional organization of mammalian retina},
  author={Kuffler, Stephen W},
  journal={Journal of neurophysiology},
  volume={16},
  number={1},
  pages={37--68},
  year={1953}
}

@article{barlow1953summation,
  title={Summation and inhibition in the frog's retina},
  author={Barlow, Horace B},
  journal={The Journal of physiology},
  volume={119},
  number={1},
  pages={69},
  year={1953}
}

@article{mcgrath2023hydra,
  title={The hydra effect: Emergent self-repair in language model computations},
  author={McGrath, Thomas and Rahtz, Matthew and Kramar, Janos and Mikulik, Vladimir and Legg, Shane},
  journal={arXiv preprint arXiv:2307.15771},
  year={2023}
}

@article{goodale1992separate,
  title={Separate visual pathways for perception and action},
  author={Goodale, Melvyn A and Milner, A David},
  journal={Trends in neurosciences},
  volume={15},
  number={1},
  pages={20--25},
  year={1992},
  publisher={Elsevier}
}

@article{solomon2004profound,
  title={Profound contrast adaptation early in the visual pathway},
  author={Solomon, Samuel G and Peirce, Jonathan W and Dhruv, Neel T and Lennie, Peter},
  journal={Neuron},
  volume={42},
  number={1},
  pages={155--162},
  year={2004},
  publisher={Elsevier}
}

@incollection{Ungerleider1982,
  author    = {Ungerleider, Leslie G. and Mishkin, Mortimer},
  title     = {Two cortical visual systems},
  booktitle = {Analysis of Visual Behavior},
  editor    = {Ingle, David J. and Goodale, Melvyn A. and Mansfield, Richard J. W.},
  publisher = {MIT Press},
  address   = {Cambridge, MA},
  year      = {1982},
  pages     = {549--586}
}

@article{kriegeskorte2015deep,
  title={Deep neural networks: a new framework for modeling biological vision and brain information processing},
  author={Kriegeskorte, Nikolaus},
  journal={Annual review of vision science},
  volume={1},
  pages={417--446},
  year={2015},
  publisher={Annual Reviews}
}

@article{nassi2009parallel,
  title={Parallel processing strategies of the primate visual system},
  author={Nassi, Jonathan J and Callaway, Edward M},
  journal={Nature reviews neuroscience},
  volume={10},
  number={5},
  pages={360--372},
  year={2009},
  publisher={Nature Publishing Group UK London}
}

@article{kaplan1986primate,
  title={The primate retina contains two types of ganglion cells, with high and low contrast sensitivity.},
  author={Kaplan, Ehud and Shapley, Robert M},
  journal={Proceedings of the National Academy of Sciences},
  volume={83},
  number={8},
  pages={2755--2757},
  year={1986}
}

@article{derrington1984spatial,
  title={Spatial and temporal contrast sensitivities of neurones in lateral geniculate nucleus of macaque.},
  author={Derrington, AM and Lennie, P},
  journal={The Journal of physiology},
  volume={357},
  number={1},
  pages={219--240},
  year={1984},
  publisher={Wiley Online Library}
}

@article{richards2019deep,
  title={A deep learning framework for neuroscience},
  author={Richards, Blake A and Lillicrap, Timothy P and Beaudoin, Philippe and Bengio, Yoshua and Bogacz, Rafal and Christensen, Amelia and Clopath, Claudia and Costa, Rui Ponte and de Berker, Archy and Ganguli, Surya and others},
  journal={Nature neuroscience},
  volume={22},
  number={11},
  pages={1761--1770},
  year={2019},
  publisher={Nature Publishing Group US New York}
}

@article{hinton2015distilling,
  title={Distilling the knowledge in a neural network},
  author={Hinton, Geoffrey and Vinyals, Oriol and Dean, Jeff},
  journal={arXiv preprint arXiv:1503.02531},
  year={2015}
}

@article{bereska2024mechanistic,
  title={Mechanistic interpretability for AI safety--a review},
  author={Bereska, Leonard and Gavves, Efstratios},
  journal={arXiv preprint arXiv:2404.14082},
  year={2024}
}

@article{dicarlo2012does,
  title={How does the brain solve visual object recognition?},
  author={DiCarlo, James J and Zoccolan, Davide and Rust, Nicole C},
  journal={Neuron},
  volume={73},
  number={3},
  pages={415--434},
  year={2012},
  publisher={Elsevier}
}

@article{lecun2015deep,
  title={Deep learning},
  author={LeCun, Yann and Bengio, Yoshua and Hinton, Geoffrey},
  journal={nature},
  volume={521},
  number={7553},
  pages={436--444},
  year={2015},
  publisher={Nature Publishing Group UK London}
}

@article{isaacson2011inhibition,
  title={How inhibition shapes cortical activity},
  author={Isaacson, Jeffry S and Scanziani, Massimo},
  journal={Neuron},
  volume={72},
  number={2},
  pages={231--243},
  year={2011},
  publisher={Elsevier}
}

@inproceedings{rozemberczkiShapleyValueMachine2022,
  title = {The {{Shapley Value}} in {{Machine Learning}}},
  booktitle = {Proceedings of the {{Thirty-First International Joint Conference}} on {{Artificial Intelligence}}},
  author = {Rozemberczki, Benedek and Watson, Lauren and Bayer, P{\'e}ter and Yang, Hao-Tsung and Kiss, Oliv{\'e}r and Nilsson, Sebastian and Sarkar, Rik},
  year = 2022,
  pages = {5572--5579},
  publisher = {International Joint Conferences on Artificial Intelligence Organization},
  doi = {10.24963/ijcai.2022/778},
  urldate = {2026-11-17},
  langid = {american}
}

@article{owenMultilinearExtensionsGames1972,
  title = {Multilinear {{Extensions}} of {{Games}}},
  author = {Owen, Guillermo},
  year = 1972,
  journal = {Management Science},
  volume = {18},
  number = {5},
  eprint = {2661445},
  eprinttype = {jstor},
  pages = {P64-P79},
  publisher = {INFORMS},
  issn = {0025-1909},
  urldate = {2026-11-17}
}

@inproceedings{kornblith2019similarity,
  title     = {Similarity of Neural Network Representations Revisited},
  author    = {Kornblith, Simon and Norouzi, Mohammad and Lee, Honglak and Hinton, Geoffrey},
  booktitle = {Proceedings of the 36th International Conference on Machine Learning},
  series    = {Proceedings of Machine Learning Research},
  volume    = {97},
  pages     = {3519--3529},
  publisher = {PMLR},
  year      = {2019}
}
